\documentclass[preprint, 12pt]{elsarticle}

\usepackage[utf8]{inputenc}
\usepackage{amssymb}
\usepackage{amsmath}
\usepackage{mathtools}
\usepackage{bm}

\let\labelindent\relax
\usepackage{enumitem}
\usepackage{dsfont}
\usepackage{graphicx}
\usepackage{float}
\usepackage[linesnumbered,lined,ruled]{algorithm2e}
\usepackage{pifont}
\usepackage{subfiles}
\usepackage{subcaption}
\usepackage{xcolor} 
\usepackage{verbatim}
\usepackage{balance}

\usepackage{hyperref}

\date{May 2023}

\begin{document}
\begin{frontmatter}

\title{Leveraging Randomized Smoothing \\ for Optimal Control of Nonsmooth Dynamical Systems } 

\author[inria]{Quentin Le Lidec\corref{cor1}}
\cortext[cor1]{Corresponding author.}
\ead{quentin.le-lidec@inria.fr}
\author[inria]{Fabian Schramm}
\author[inria,prague]{Louis Montaut}
\author[inria]{Cordelia Schmid}
\author[inria]{Ivan Laptev}
\author[inria]{Justin Carpentier}

\address[inria]{Inria and Département d'Informatique de l'Ecole Normale Supérieure, PSL Research University, Paris, France}
\address[prague]{Czech Institute of Informatics, Robotics and Cybernetics, Czech Technical University, Prague, Czech Republic}

\begin{abstract}
    Optimal control (OC) algorithms such as differential dynamic programming (DDP) take advantage of the derivatives of the dynamics to control physical systems efficiently. 
    Yet, these algorithms are prone to failure when dealing with non-smooth dynamical systems. This can be attributed to factors such as the existence of discontinuities in the dynamics derivatives or the presence of non-informative gradients.
    On the contrary, reinforcement learning (RL) algorithms have shown better empirical results in scenarios exhibiting non-smooth effects (contacts, frictions, etc.). 
    Our approach leverages recent works on randomized smoothing (RS) to tackle non-smoothness issues commonly encountered in optimal control and provides key insights on the interplay between RL and OC through the prism of RS methods. 
    This naturally leads us to introduce the randomized Differential Dynamic Programming (RDDP) algorithm accounting for deterministic but non-smooth dynamics in a very sample-efficient way. 
    The experiments demonstrate that our method can solve classic robotic problems with dry friction and frictional contacts, where classical OC algorithms are likely to fail, and RL algorithms require, in practice, a prohibitive number of samples to find an optimal solution.
\end{abstract}


\end{frontmatter}

\section{Introduction}
Theories and applications of optimal control (OC) and reinforcement learning (RL) are all related to the problem of minimizing a cost (resp. maximizing a reward) while fulfilling the system dynamics and constraints over a given time duration.
Nonetheless, the resulting algorithms to solve OC or RL problems are based on different approaches, leading to very different performances in practice.
On the one hand, in their vast majority, RL algorithms only exploit samples, leading to zero-th order approaches.
On the other hand, optimal control and trajectory optimization techniques such as the iterative Linear Quadratic Regulator (iLQR)~\cite{li2004iterative} and Differential Dynamic Programming (DDP)~\cite{tassa2012synthesis} rely on first-order and second-order linearization of the dynamics. 
Exploiting this derivative information makes them much more sample-efficient than their zero-th order counterparts from the field of RL.
However, when considering complex scenarios such as robots interacting with their environments, the system dynamics may depict some non-smooth physical phenomena (dry friction, contact constraints, etc.). 
These properties may induce non-informative or discontinuous gradients that make gradient-based strategies fail~\cite{werling2021fast}.
On the contrary, RL algorithms have proven to be able to get around these non-smoothness issues in such cases, leading to impressive results when considering contact interactions~\cite{hwangbo2019learning}. 
By treating the dynamics as a black-box function, derivative-free algorithms such as standard RL methods circumvent the issues as mentioned above. They can transparently deal with arbitrarily complex and non-smooth dynamics. 
However, completely disregarding the specific structure of the dynamics comes at the cost of often requiring a large number of samples.

\begin{figure}[t]
    \centering
    \includegraphics[width=0.60 \linewidth]{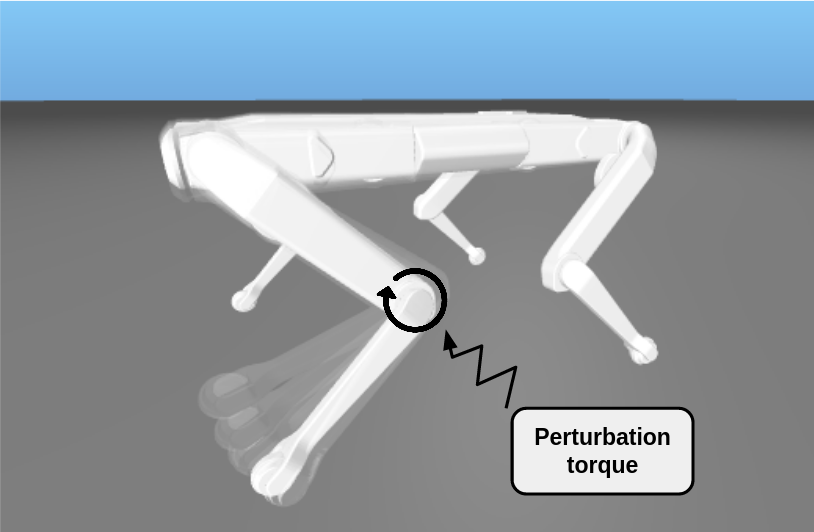}
    \caption{Illustration of randomized smoothing effects on the front left leg of the Solo robot.}
    \label{fig:target_solo}
    \vspace{-0.5cm}
\end{figure}

In a recent growing effort, differentiable physical simulators have emerged in the context of exploiting informative gradients for control~\cite{NEURIPS2018_842424a1} and estimation~\cite{le2021differentiable}.
Gradients of rigid body dynamics were obtained by differentiating the classical rigid body algorithms~\cite{featherstone2014rigid, carpentier2018analytical}. 
Simulating and differentiating physics with frictional contacts is more challenging as it requires to also solve for contact forces~\cite{NEURIPS2018_842424a1,le2021differentiable} and was made possible by differentiable optimization techniques~\cite{amos2017optnet,agrawal2019differentiable}. 
Yet, these dynamics derivatives may present some discontinuities or lack of regularity, drastically impacting gradient-based optimization techniques, especially in the context of classical control algorithms~\cite{werling2021fast}.

In this work, we propose to leverage randomized smoothing (RS) techniques \cite{duchi2012randomized,NEURIPS2020_6bb56208} to cope with nonsmooth dynamical systems in optimal control problems.
From a theoretical perspective, we notably demonstrate how RS methods applied to OC problems allow us to close the gap with the RL setting (Sec.~\ref{sec:bridging_the_gap}) in general.
A first connection of RL and OC for control tasks with infinite state and action spaces has been introduced by  \cite{bradtke1992reinforcement, bradtke1994adaptive} for the linear quadratic regulator with a focus on convergence proofs.
From a practical perspective, we propose to use RS techniques within the frame of Differential Dynamic Programming to deal with nonsmooth dynamics, which are hard to handle in the vanilla setting (Sec.~\ref{sec:RDDP}).
We also introduce an adaptive strategy to automatically reduce the smoothing noise inherent to RS techniques across the optimization procedure.
We experimentally show the practical benefits of our approach on various robotic tasks with increasing complexity, ranging from inverted pendulum to locomotion (Sec.~\ref{sec:exp}).

Similarly to our work, \cite{suh2021bundled} introduced the notion of randomized smoothing on the state and control spaces to get gradients from the dynamics through contacts.
The same authors extended the approach to motion planning for complex manipulation tasks \cite{pang2022global} by exploring the influence of zero-th and first-order gradient estimators.
We take a different point of view by establishing links between RL and OC through the lens of RS, which is a first step towards a stronger interplay between these two fields. 
This viewpoint implies a practical difference: we interpret RS as an exploration term and only smooth the control space, leading to a sampled space of reduced dimension.
Our experiments demonstrate that this is sufficient to perform well on complex robotics tasks.

\section{Background}\label{sec:background}

Our work builds on optimal control, reinforcement learning, and randomized smoothing techniques applied to robotics, which are briefly introduced in this section.\\

\noindent{\textbf{Optimal control.}} 
\label{sec:background_DDP}
We consider the OC problem of controlling a robot by minimizing a cost $l$ while satisfying the system dynamics $f$:
\begin{subequations}
    \begin{alignat}{2}
        \min_{x,u} &  \int_0^T l(\tau,x(\tau),u(\tau))\text{d}\tau  \\
        \text{s.t.} & \ \Dot{x}(\tau) = f(x(\tau),u(\tau)), \\
                    & \ x(0) = \hat{x}_0,
    \end{alignat}
    \label{eq:continuous_OC}
\end{subequations}
where $x(\tau) \in \mathcal{S}$ and $u(\tau) \in \mathcal{A}$ are the state and the control action of the system at time $\tau$ respectively, $\hat{x}_0 \in \mathcal{S}$ is a given initial state and $T$ is the time horizon.
In this paper, we call \textit{smooth} a dynamics whose corresponding function $f$ is differentiable everywhere.
While in robotics $f$ is often considered smooth, contact interactions give rise to points where $f$ is only sub-differentiable~\cite{brogliato1999nonsmooth}.

Two different approaches may be used to solve \eqref{eq:continuous_OC}, namely \textit{direct} and \textit{indirect} approaches.
\textit{Direct} approaches translate the infinite dimensional problem~\eqref{eq:continuous_OC} into a finite nonlinear programming problem~(NLP) and exploit off-the-shelf constrained optimization solvers \cite{diehl2006fast} for solving it. In particular, this leads to a discrete numerical problem of the form:
\begin{subequations}
    \begin{alignat}{2}
  \min_{\bm{x},\bm{u}} & \quad \overset{J(\boldsymbol{x},\boldsymbol{u})}{ \overbrace{l_N(x_N) + \sum_{t = 0}^{N-1} l_t(x_t,u_t)  }}  \label{eq:OC_problem_cost}\\
  \text{s.t.} &   \ x_{t+1} = f(x_t,u_t),\ \forall t \in \left[0,N-1\right], \label{eq:dyn_cons} \\
& x_0 = \hat{x}_0,
    \end{alignat}
    \label{eq:OC_problem}
\end{subequations}
where \mbox{$\boldsymbol{x} = \{ x_0, \dots, x_N \}$} and \mbox{$\boldsymbol{u} = \{ u_0, \dots, u_{N-1}  \}$}, with $N$ being the number of time steps. 

Alternatively, \textit{indirect} approaches first apply the optimality conditions of optimal control problems (e.g., Hamilton-Jacobi-Bellman or Pontryagin's maximum principles) and then discretize these conditions in order to solve the problems numerically.
This notably offers the advantage of highlighting the underlying sparsity of constraints induced by times. 
In this line of work, iLQR~\cite{li2004iterative} and DDP~\cite{mayne1966second,tassa2012synthesis,mastalli2020crocoddyl} are the most well-known first and second-order algorithms, with linear or quadratic-type convergence rates respectively. 
Moreover, their recent adaptations can even handle trajectory constraints~\cite{howell_altro_2019,kazdadi2021equality, jallet:hal-04332348} and implicit dynamics~\cite{jalletImplicitDifferentialDynamic2022}. 
More closely related to our work, sampled DDP \cite{sampled_ddp} lies in between, as it uses stochastic estimates of the dynamics derivatives to deal with dynamics for which gradients are unavailable.

\noindent{\textbf{Reinforcement learning}} considers the dynamics as a black-box function and uses parameterized stochastic policies $\pi_\theta$ to explore the action space better. 
This model-free approach can naturally handle complex or even unknown dynamics.
During training, the cumulated sum of rewards R along the trajectories sampled from the distribution $\rho_\theta$ induced by the policy $\pi_\theta$ is maximized 

the  along the trajectories sampled from the distribution $\rho_\theta$ induced by the policy $\pi_\theta$ is minimized (which is equivalent to minimize the average cost $J=-R$), leading to the following problem:   
\begin{align}
    \max_{\theta}& \ \mathbb{E}_{(\boldsymbol{x},\boldsymbol{u})\sim \rho_\theta} \left[ R(\boldsymbol{x},\boldsymbol{u}) \right]
    \label{eq:uncons_policyRL}
\end{align}
where trajectories are generated by the policy in the following way:
\begin{subequations}
    \begin{alignat}{2}
    u_t & \sim \pi_\theta(\cdot | x_t) , \\
    x_{t+1} &= f(x_t,u_t).
    \end{alignat}
\end{subequations}
Policy Gradient (PG)~\cite{williams1992simple, sutton2018reinforcement} algorithms aim at maximizing \eqref{eq:uncons_policyRL} by using a zero-th order estimate of the gradient as a ascent direction:
\begin{align}
    \nabla_\theta R_{PG} &= \mathbb{E}_{(\boldsymbol{x},\boldsymbol{u})\sim \rho_\theta} \left[ R(\boldsymbol{x},\boldsymbol{u}) \nabla_\theta \log \rho_\theta(\boldsymbol{x}, \boldsymbol{u})\right]  \label{eq:grad_PG}
\end{align}

The resulting randomness induces some variance in the gradient estimates, which slows down optimization and fosters exploratory behaviors, potentially leading to more global solutions.
Finally, this makes RL an instance of random optimization, which will be discussed in more detail in Sec.~\ref{sec:rs_rl}.

\noindent{\textbf{Random optimization}} 
\label{sec:background_RS} 
techniques are among the earliest optimization schemes and were developed in order to tackle problems where the objective function is discontinuous or nonsmooth.
In this situation, the gradients only provide limited information and are not suited for use in classic gradient-based optimization techniques~\cite{matyas1965random}.
During a random search, one classical strategy consists in sampling a random direction at each step and then moving towards the corresponding directional derivative~\cite{polyak1987randomopt}, which is the basis of the Simultaneous Perturbation Stochastic Approximation algorithm \cite{spall2005introduction}.
Gradient Sampling (GS) is an alternative strategy that approximates the sub-gradient at non-differentiable points by taking the descent direction inside the convex hull of some gradients randomly sampled in a neighborhood~\cite{burke2005robust}.
Recent works~\cite{nesterov2017random} exploit the equivalence between random optimization techniques and perform stochastic gradient descent on a smoothed version of the original problem to get theoretical convergence bounds.
In a parallel line of work, randomized smoothing was recently introduced in the machine learning community in order to be able to differentiate through Linear Programming (LP) \textit{argmin} operators \cite{abernethy2016perturbation,NEURIPS2020_6bb56208}.
It was also applied to make rendering \cite{lelidec:hal-03378451}, patch selection\cite{cordonnier2021differentiable}, clustering \cite{stewart2023differentiable} and collision detection \cite{montaut2022differentiable} operations differentiable.
Unlike their classical counterparts, these perturbed optimizers are guaranteed to have non-null gradients everywhere, making it possible to use gradient-based optimization methods.

Concretely, let $Z$ be a random variable whose probability distribution $\mu$ is a Gibbs measure. 
A function $g$ can be approximated by convolving it with this probability distribution:
\begin{equation}
    g_\epsilon(x) = \mathbb{E}_{Z\sim \mu} \left[ g(x + \epsilon Z) \right]
\end{equation}
which corresponds to the randomly smoothed counterpart of $g$ and can be estimated with a Monte-Carlo estimator as follows:
\begin{equation}
    g_\epsilon(x) \approx \frac{1}{M} \sum_{i=0}^M g(x + \epsilon Z^{(i)})
\end{equation}
where $\{ Z^{(1)}, \dots, Z^{(M)}\}$ are i.i.d. samples and $M$ is the number of samples.
Using integration by part, we have the following expression of gradients:
\begin{subequations}
    \begin{alignat}{2}
        \nabla_x g_\epsilon(x) &= \mathbb{E}_{Z \sim \mu} \left[ -g(x + \epsilon Z) \frac{\nabla \log \mu (Z)^\top}{\epsilon} \right] \label{eq:zero_RS}\\
     &=  \mathbb{E}_{Z\sim \mu} \left[ \nabla g(x + \epsilon Z) \right], \label{eq:first_RS}
        \end{alignat}
\end{subequations}
where \eqref{eq:zero_RS} and \eqref{eq:first_RS} corresponds respectively to the zero-th and first order expressions of $\nabla_x g_\epsilon$.
In practice, it is possible to approximate the smoothed gradient by the first order Monte-Carlo (MC) estimator~\eqref{eq:MC_first_order} or, because \mbox{$\mathbb{E}_{Z\sim \mu} \left[ \nabla \log \mu (Z)^\top \right] = 0$}, by the variance reduced zero-th order MC estimator~\eqref{eq:MC_RS}:
\begin{subequations}
    \begin{alignat}{2}
        \nabla_x g_\epsilon (x) 
        & \approx \frac{1}{M} \sum_{i=1}^M \left(g(x) -g(x + \epsilon Z^{(i)}) \right)\frac{\nabla \log \mu (Z^{(i)})^\top}{\epsilon} \label{eq:MC_RS} \\
        & \approx \frac{1}{M} \sum_{i=1}^M \nabla g(x + \epsilon Z^{(i)}) \label{eq:MC_first_order}
        \end{alignat}
\end{subequations}
More intuitively, because of the local averaging effect of the convolution, $g_\epsilon(x)$ is always smoother than $g$.
Indeed, $g_\epsilon$ is guaranteed to be differentiable \cite{bertsekas1973stochastic}, uniformly close from $g$ and its gradient to be Lipschitz-continuous \cite{duchi2012randomized,NEURIPS2020_6bb56208}. 
Moreover, $g_\epsilon \xrightarrow{\epsilon \xrightarrow[]{}0} g$, so reducing the perturbation by decreasing $\epsilon$ leads to a reduced gap between $g_\epsilon$ and the original function $g$ but also results in a less smooth approximation.
In terms of computational complexity, evaluating $\nabla g_\epsilon$ with $M$ samples induces a complexity increased by a factor $M$.
The computation being easily parallelizable, this, in fact, leads to a constant computational time without a critical impact on the memory footprint, as shown in the context of differentiable rendering in~\cite{lelidec:hal-03378451}. 
Adding stochasticity to gradients is also proven to help escape saddle points when optimizing non-convex functions \cite{ge2015escaping}, which constitutes a positive side effect of randomized smoothing.
Finally, other previous works on the use of randomized smoothing demonstrate how it can improve convergence rates when optimizing non-smooth functions~\cite{duchi2012randomized} and lead to more robust solutions~\cite{cohen2019certified}. 

\noindent{\textbf{Non-smooth dynamics in robotics.}} Generating movements for locomotion and manipulation is considered a major task in robotics.
Whether it is to move itself or an object, the robot has to interact with its environment by making and breaking contacts, which induce non-smooth dynamics \cite{BROGLIATO2023297}.
It is common to address these issues by making the hypothesis of bilateral contacts either between feet and the ground for locomotion \cite{farshidian2017efficient} or between the fingers and the manipulated object \cite{mason1985robot} as detailed in Appendix~\ref{sec:app-ncp}.
This hypothesis limits the range of possible movements by enabling only conservative strategies and fixing the contact modes during predetermined phases.
RL approaches go beyond these constraints by not making any assumption on the underlying physical model and, hence, are said to be model-free.
To go towards more dynamical movements, model-based approaches should be able to handle the switches between various contact modes better \cite{carpentier2021recent, wensing2023optimization}.
Direct optimization of the sequence of contacts \cite{posa2014direct,toussaint2018differentiable} avoids the tedious manual engineering by shifting the burden to mixed-integer solvers. 
Alternatively, the seminal work from \cite{mordatch2012discovery} exploits an analytical smoothing of the dynamics to use classical smooth optimization techniques.
This work and the concurrent \cite{suh2021bundled} propose to apply randomized smoothing on the dynamics before leveraging Trajectory Optimization algorithms.

\begin{figure}[t]
    \centering
    \includegraphics[width=0.95 \linewidth]{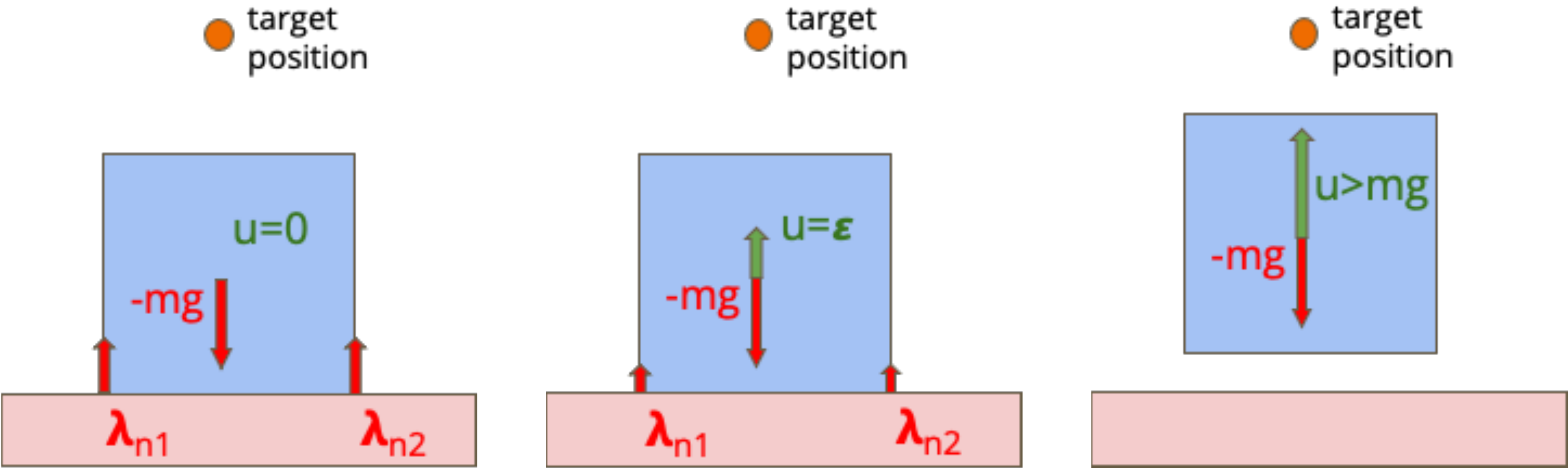}\\[0.5cm]
    \includegraphics[width=0.7 \linewidth]{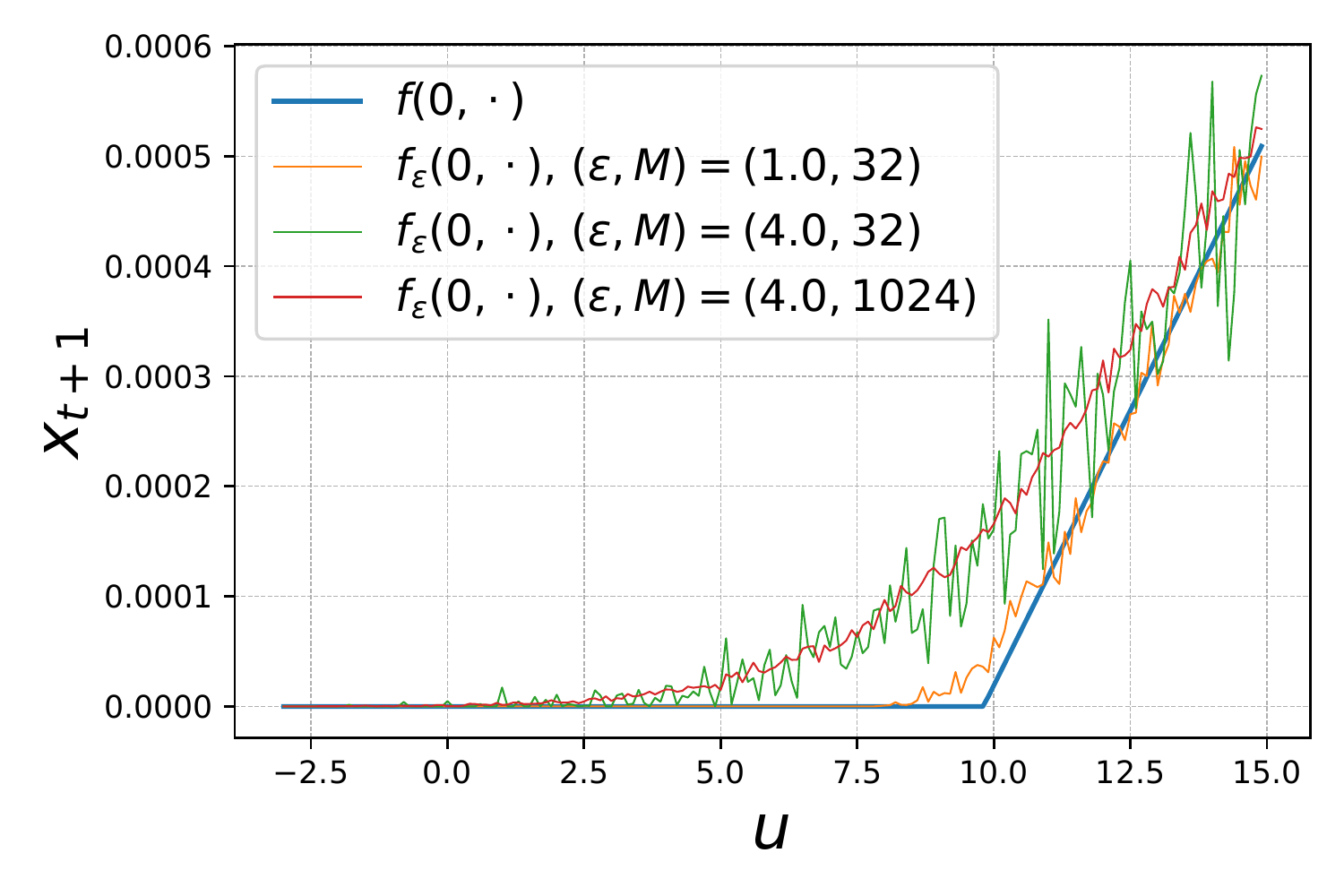}
    \caption{\textbf{Top:} A slight vertical force cannot break the unilateral contact, leaving the cube on the floor and, thus, the state unchanged. \textbf{Bottom:} The non-smoothness of physics induces null gradients $\nabla_{\boldsymbol{u}} J$ which results in the failure of classical optimization techniques (\ref{sec:local_OC}).}
    \label{fig:cube_schema}
    \vspace{-0.5cm}
\end{figure}

\section{Bridging the gap between optimal control and reinforcement learning} \label{sec:bridging_the_gap}

In this section, we present the caveats of poorly informative gradients for classical control algorithms, which may, for instance, occur in the presence of nonsmooth dynamical systems.
To overcome these limitations, we propose to exploit the randomized smoothing approach in the trajectory optimization paradigm.
We also detail a connection between randomized smoothing and RL methods, thus explaining how they effectively solve problems involving poorly informative gradients. 

\subsection{Locally optimal solutions of optimal control problems} 
\label{sec:local_OC}

As discussed in Sec.~\ref{sec:background_DDP}, several approaches may be used for solving OC problems of the form~\eqref{eq:OC_problem}.
In particular, one can substitute $x_1, \dots, x_N$ thanks to the constraint on the dynamics~\eqref{eq:dyn_cons} and express problem~\eqref{eq:OC_problem} only in terms of $u_0, \dots, u_{N-1}$, leading to the following but equivalent unconstrained optimization problem:
\begin{equation}
    \min_{\boldsymbol{u}} \, J(\boldsymbol{x}(\boldsymbol{u}),\boldsymbol{u})
    \label{eq:uncons_OC}
\end{equation}
where $\boldsymbol{x}(\boldsymbol{u})$ is recursively defined by:
\begin{equation}
\label{eq:dyn_cons}
    x_0 = \hat{x}_0 ~\text{and}~ x_t(\boldsymbol{u}) = f(x_{t-1}(\boldsymbol{u}),u_t),
\end{equation}
corresponding to an integration process (e.g., exploiting a dynamical simulator).
Unrolling the successive integration steps facilitates the efficient differentiation of $\boldsymbol{x}(\boldsymbol{u})$.
Solving the equivalent problem~\eqref{eq:uncons_OC} can be achieved using a classical unconstrained optimization algorithm such as gradient descent, consisting of backpropagating through time~\cite{bengio1994learning}. 
In the case of a robotic system, the dynamic simulator involves making and breaking contacts and switching between sticking and sliding contacts, which induces several modes in the dynamics.
More formally, the computation of $f$ results from a Non-linear Complementarity Problem (NCP), making it inherently non-smooth as detailed in the Appendix~\ref{sec:app-ncp} and \cite{acary2017contact, lidec2023contact}.
In addition, the cost function $J$, which often involves the robot kinematics via a penalty on the end-effector position as done in the experiments of Sec.~\ref{sec:exp}, is non-convex.
Thus, solving~\eqref{eq:OC_problem} with approaches described previously can lead to local solutions because of the inherent non-convexity and non-smoothness of the problem.

To illustrate this, one can think about the problem of lifting a cube~\cite{werling2021fast} illustrated in Fig.~\ref{fig:cube_schema}.
When $\boldsymbol{u}$ is initialized with null control, this results in $ \nabla_{\boldsymbol{u}} f =0$ because of the complementarity constraint arising from unilateral contacts (see IV. of \cite{werling2021fast}).
By supposing that $\frac{\partial J}{\partial \boldsymbol{u}}=0$ and applying the chain rule, we have that:
\begin{equation}
    \nabla_{\boldsymbol{u}}J(\boldsymbol{x}(\boldsymbol{u}), \boldsymbol{u}) 
    = \frac{\partial J}{\partial \boldsymbol{x}}\frac{\partial \boldsymbol{x}}{\partial \boldsymbol{u}} + \frac{\partial J}{\partial \boldsymbol{u}}
    =  \frac{\partial J}{\partial \boldsymbol{x}}\frac{\partial \boldsymbol{x}}{\partial \boldsymbol{u}}.
\end{equation}
Moreover, since $\nabla_{\boldsymbol{u}} f =0$, we have that starting from the first time step \mbox{$\frac{\partial x_1}{\partial \boldsymbol{u}}= \frac{\partial f}{\partial \boldsymbol{u}}(x_0,u_0) = 0$} and with the time recursion, this leads to: 
\begin{align}
    \frac{\partial x_{t+1}}{\partial \boldsymbol{u}}= \frac{\partial f}{\partial \boldsymbol{u}}(x_t,u_t) + \frac{\partial f}{\partial \boldsymbol{x}}(x_t,u_t)\frac{\partial x_t}{\partial \boldsymbol{u}} = 0,
\end{align}
implying that $\frac{\partial \boldsymbol{x}}{\partial \boldsymbol{u}} = 0$. 
Finally, because $\nabla_{\boldsymbol{u}} J = 0$, an algorithm exploiting only the local gradient information (\textit{e.g.,} gradient or Newton descent) will stop at this point, leaving the problem unsolved and blocked at a local maximum. The classical DDP algorithm would get stuck in a similar situation for the same reasons. 
It is also worth mentioning that even when equipped with proximal smoothing \cite{parikh2014proximal}, such algorithms would still rely on local information and, thus, fail to overcome the issue of non-informative gradients.

\subsection{Randomized smoothing of the system dynamics}

The issue highlighted above is due to the inability of deterministic control algorithms to deal with non-smooth dynamics and their non-informative gradients, which often occur for physical systems involving contact or friction.

An intuitive way to circumvent this issue consists of introducing randomization in the optimization process in order to get a more exploratory behavior by collecting samples in a larger neighborhood around a given point when compared to classic methods such as the ones relying on local gradient information. 
This is precisely the motivation behind Randomized Smoothing and, to some extent, behind Reinforcement Learning, as discussed in Sec.~\ref{sec:rs_rl}.
We adapt the formulation \eqref{eq:OC_problem} by artificially smoothing the system dynamics using randomized smoothing, leading to the following smooth but approximated problem:
\begin{subequations}
    \begin{alignat}{2}
  \min_{\boldsymbol{x},\boldsymbol{u}}& \ J(\boldsymbol{x},\boldsymbol{u})  \\
  \text{s.t.} &   \ x_{t+1} = f_\epsilon (x_t,u_t),\ \forall t \in \left[0,N-1\right], \label{eq:smooth_dyn} \\
 & x_0 = \hat{x}_0.
    \end{alignat}
    \label{eq:smooth_OC_problem}
\end{subequations}
where $ f_\epsilon(x,u) = \mathbb{E}_{Z\sim \mu}\left[ f(x,u + \epsilon Z)\right]$ and $\mu$ is a noise distribution.
It is worth mentioning that $f_\epsilon$ corresponds to a randomized version of $f$, which is only perturbed with respect to the control input $u$.
Perturbing the control but not the state ensures that only reachable states are explored.
When $\epsilon \rightarrow 0$, problem~\eqref{eq:smooth_OC_problem} converges to the original problem~\eqref{eq:OC_problem}.

The proposed solution of using a smoothed approximation of the system dynamics is very generic. It can be easily instantiated in most of the existing trajectory optimization frameworks without major modifications.
Yet, there is \textit{a priori} no obvious choice of the sampling distribution $\mu$, neither for the number of particles sufficient in the Monte-Carlo estimator of the system dynamics and gradient computations.
Another difficulty lies in the proper scheduling of the noise intensity $\epsilon$ towards $0$ in order to remove the effect of noise at convergence to recover the original problem.

In Sec.~\ref{sec:RDDP}, we introduce an algorithmic variation of the so-called Differential Dynamic Programming algorithm, which relies on the smoothed dynamics~$f_\epsilon$. In particular, we propose an automatic scheduling of the noise intensity~$\epsilon$ and an auto-tuning strategy of the number of particles in the Monte-Carlo estimators.

\subsection{Reinforcement learning through the prism of randomized smoothing} 
\label{sec:rs_rl}

At this stage, one could wonder why RL demonstrated empirical success even in the case of the non-smooth dynamics evoked in \ref{sec:local_OC}. We provide the first possible explanation by drawing a parallel between the descent directions used in RL and the one from random optimization \eqref{eq:zero_RS} when applied to the OC problem \eqref{eq:OC_problem}.

Indeed, the ascent directions used in the classical RL algorithm REINFORCE with baseline (the same considerations remain valid for the closely related actor-critic algorithms) \cite{greensmith2004variance,sutton2018reinforcement, mnih2016asynchronous} can be written as:
\begin{equation}
    \nabla_\theta R_{PG} = \mathbb{E}_{(\boldsymbol{x},\boldsymbol{u})\sim \rho_\theta} \Bigg[ \left(R(\boldsymbol{x},\boldsymbol{u}) -\hat{V}(\hat{x}_0) \right) \nabla_\theta \log \rho_\theta(\boldsymbol{x},\boldsymbol{u})\Bigg],  
    \label{eq:reinforce_with_baseline}
\end{equation}
where $\hat{V}$ is an estimate of the value function, which exactly corresponds to the variance-reduced version of the randomly smoothed approximation \eqref{eq:MC_RS}.
More intuitively, in order to deal with the non-smoothness of the dynamics or reward functions, Policy Gradient adds noise in the action space by sampling actions from a stochastic policy. 
Concretely, this causes $\nabla_\theta R_{PG} \neq 0 $ even in regions where $\nabla_{\boldsymbol{u}} R = 0$ and first or second order gradient methods fail, as discussed in Sec.~\ref{sec:local_OC}.
Thus, introducing some stochasticity allows RL to smooth the original problem and avoid the computation of gradients from the dynamics when they are unknown.
Unfortunately, this generally comes at the cost of increased variance in the estimates of $\nabla R_{PG}$, which induces a slower convergence rate~\cite{nesterov2017random}.
A similar study could be done with Evolution Strategies used in~\cite{salimans2017evolution,mania2018simple}, which prefer to generate trajectories with deterministic policies but parameterized by randomly sampled parameters.

Following this analysis, in a way similar to RL, we propose using randomized smoothing to compute informative gradients even when the standard ones are non-informative.  
The obtained gradients can then be exploited in the context of optimal control problems with higher-order optimization algorithms in order to get improved convergence rates, as shown in the context of the widely used Differential Dynamic Programming algorithm in the following section.

In earlier works, \cite{bradtke1992reinforcement, bradtke1994adaptive} derive a convergence proof for policy iteration methods in a deterministic, linear, and time-variant LQR setup for infinite state and action spaces. An algorithm was presented, that converges to the optimal policy and involves some additional exploration noise in the control due to a requirement in the underlying recursive least-square algorithm. However, the noise has a very different motivation from our usage of injected noise for randomized smoothing of non-smooth function surfaces, which is done by averaging over several noise samples.
\section{Randomized Differential Dynamic Programming}
\label{sec:RDDP}

This section introduces our randomized Differential Dynamic Programming algorithm. 
This novel formulation builds on the previous analysis to incorporate randomized smoothing in the optimal control paradigm in order to increase the exploration of classical DDP and efficiently solve problems involving non-smooth dynamical systems. 

\subsection{Dynamic programming with smoothed physics}

\begin{algorithm}[t]
\SetAlgoLined
  \KwIn{OC problem: $J,f$, initial trajectory: $\bm{x},\bm{u}$, target noise and precision: $ \epsilon^*, \alpha^*$, initial noise and precision: $\epsilon, \alpha$, adaptive scheme parameters: $\rho, \gamma$}
  \KwOut{Solution $\bm{x},\bm{u}$ of the OC problem \eqref{eq:OC_problem}}
 \Repeat{$\alpha < \alpha^* \ \mathrm{and} \ \epsilon < \epsilon^*$}{
    \Repeat{$\| Q_u \|_{Q_{uu}^{-1}} < \alpha$}{
        $k,K \leftarrow \textbf{Backward Pass \eqref{eq:backward_DDP}}$\;
        $\bm{x},\bm{u} \leftarrow \textbf{Forward Pass \eqref{eq:forward_DDP}}$\;
    }
    $\epsilon \leftarrow \epsilon/\rho$ \;
    $\alpha \leftarrow \alpha/\gamma$\;
 }
\caption{Randomized DDP algorithm }
\label{alg:RDDP}
\end{algorithm}

The main idea behind randomized DDP consists in exploiting the formulation \eqref{eq:smooth_OC_problem} and replacing the original, possibly non-smooth, dynamics $f$ by its randomly smoothed approximation $f_\epsilon$ in the forward and backward passes of the vanilla DDP. 
Doing so allows to benefit from the efficiency of DDP even in situations where the original DDP algorithm will fail.

First, we introduce the cost-to-go function associated to our problem \eqref{eq:smooth_OC_problem}:
\begin{align}
    J_t(x_t,u_t,\dots,u_{N-1}) = l_N(x_N) + \sum_{j = t}^{N-1} l_j(x_j,u_j)
\end{align}
and the value function which verifies the Bellman's equation:
\begin{subequations}
    \begin{alignat}{2}
    V_t(x_t) 
    &= \min_{u_t,\dots,u_{N-1}} J_t(x_t,u_t,\dots,u_{N-1}) \\
    &= \min_{u_t} \, l_t(x_t,u_t) + V_{t+1}\left(f_\epsilon(x_t,u_t)\right)  
    \end{alignat}
    \label{eq:bellman}
\end{subequations}
with the terminal condition $V_N(x) = l_N(x)$.
Additionally, the $Q$-function is defined by:
\begin{align}
    Q_t(x,u) = l_t(x,u) + V_{t+1}\left(f_\epsilon(x,u)\right) 
\end{align}

\begin{figure}[t]
    \centering
    \includegraphics[width=0.45\linewidth]{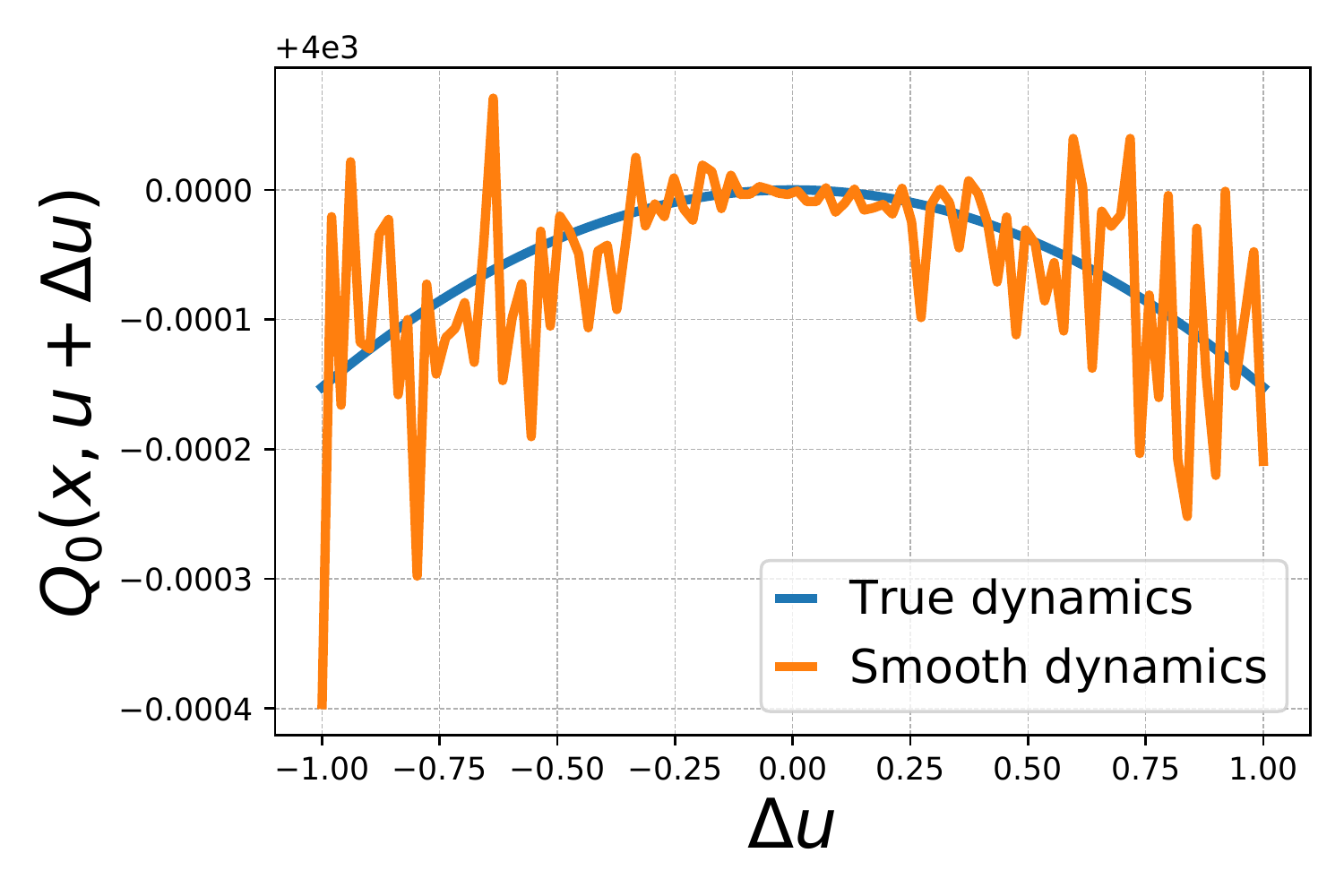}
    \includegraphics[width=0.45 \linewidth]{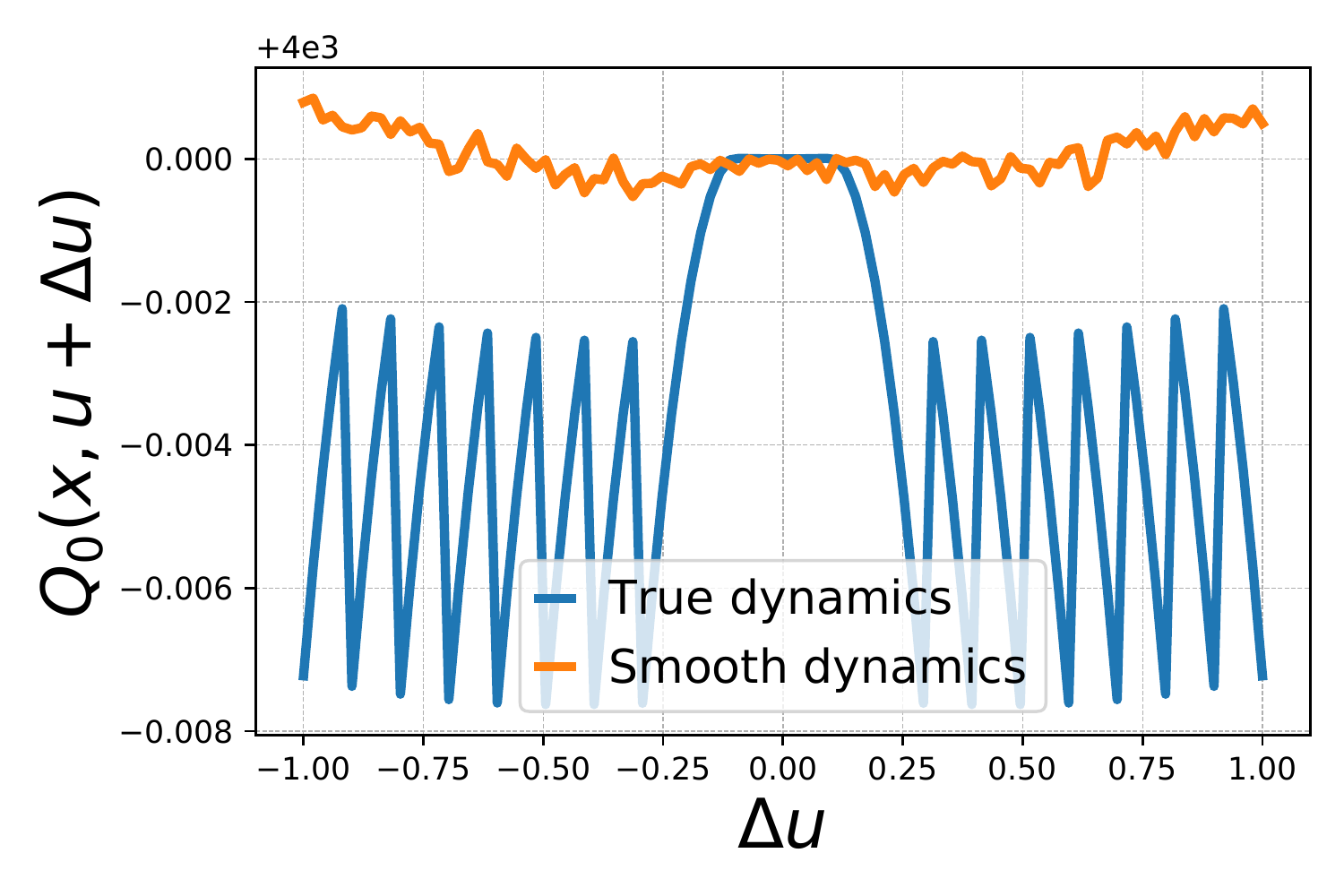}
    \caption{\textbf{Left:} The randomly-smoothed dynamics allow to get non-null gradients and escape the local maximum for the inverted pendulum. \textbf{Right:} $Q_0(x,\cdot)$ value function of the pendulum with dry friction exhibits plateaus where $Q_u=0$ around $u=0$ while randomly smoothed dynamics leads to $Q_u \neq 0$.}
    \label{fig:Qfunc_pendulum}
\end{figure}

As done in the classical Differential Dynamic Programming algorithm, we exploit the sparsity of constraints induced by time via Bellman's equation to solve the problem \eqref{eq:smooth_OC_problem}. 
To do so,  local second-order approximations of the value and the $Q$ functions are built by backpropagating the Bellman's equation \eqref{eq:bellman} backward in time around a reference trajectory $\overline{\boldsymbol{x}}$, leading to the backward pass equations:
\begin{subequations}
    \begin{alignat}{2}
    Q_{xx} &= l_{xx} + {f_x}^\top V'_{xx}f_x + {V'_x}^\top f_{xx}   \\
    Q_{ux} &= l_{ux} + {f_u}^\top V'_{xx}f_x + {V'_x}^\top f_{ux} \\
    Q_{uu} &= l_{uu} + {f_u}^\top V'_{xx}f_u + {V'_x}^\top f_{uu}  \\
    Q_x &= l_x + {V'_x}^\top f_x \\
    Q_u &= l_u + {V'_x}^\top f_u \\
    q &= l + v',
    \end{alignat}
    \label{eq:backward_DDP}
\end{subequations}
where $v = V_t(x)$, and and the subscript on $x$ and $u$ are the usual notations for the partial derivative w.r.t the state and control variables, and the superscript $V'$ denotes the value function at the next time-step.
Minimizing these local quadratic approximations of $Q$ w.r.t $u$ gives:
\begin{equation}
    u = k + K (x - \overline{x}),~ k = - Q_{uu}^{-1} Q_u ~\text{and}~ K = - Q_{uu}^{-1} Q_{ux},  
    \label{eq:opt_pol}
\end{equation}
and injecting~\eqref{eq:opt_pol} in~\eqref{eq:bellman} gives rise to:
\begin{subequations}
    \begin{alignat}{2}
    V_{xx} &= Q_{xx} - K ^\top Q_{uu} K \\
    V_x &= Q_x - k^\top Q_{uu} K \\
    v &= q - \frac{1}{2} k ^\top Q_{uu}k.
\end{alignat}
\end{subequations}
Finally, $u$ is updated with a line search during the forward computation:
\begin{subequations}
    \begin{alignat}{2}
   u^{n}_t &= u_t^{n-1} + \alpha k + K(x^{n}_t - \overline{x_t}) \\
   x^{n}_{t+1} &= f_\epsilon(x^{n}_t, u^{n}_t)
    \end{alignat}
     \label{eq:forward_DDP}
\end{subequations}
with the initial condition $x^{n}_0 = \hat{x}_0$ and where the superscript $n$ denotes the number of optimization iterations. 
During line search, the noise is fixed as described in~\cite{NEURIPS2019_2557911c, nocedal2006numerical}.
Repeating the forward and backward steps by taking the new trajectory $\boldsymbol{x}^n$ as the new reference $\overline{\boldsymbol{x}}$ allows to find the optimal control $\boldsymbol{u}$ under the form of a  linear policy \eqref{eq:opt_pol} which is optimal around the trajectory.

As detailed previously in Sec.~\ref{sec:background_RS}, smoothing the physics makes it possible to have non-null gradients $\nabla_u f_\epsilon$ thus inducing non-null $Q_u$, as illustrated in Fig.~\ref{fig:Qfunc_pendulum}.
Alternatively, the stochasticity can also be interpreted as an exploration term that has proven to be crucial in the RL framework to escape regions where the local information from gradients does not provide exploitable insight on the problem being solved (see Sec.~\ref{sec:rs_rl}). 

The main difference between our approach formulated at \eqref{eq:smooth_OC_problem} and "Policy Gradient"-type algorithms lies in the scope of the randomized smoothing. 
Indeed, RL smooths the whole problem while we "only" smooth the dynamics $f$: note that the expectation is on entire trajectories in \eqref{eq:uncons_policyRL} while it is at the time-step level in \eqref{eq:smooth_OC_problem} resulting in a reduced variance in the latter case \cite{peters2008reinforcement}.
Moreover, we find that preserving the original recursive structure of \eqref{eq:OC_problem} is beneficial in the case of known dynamics $f$ as it allows to benefit from the efficient dynamic programming backward passes \eqref{eq:backward_DDP} of DDP.
While RL requires to also smooth the cost function to deal with sparse rewards, it is not necessary in our case as current trajectory optimization algorithms can handle hard constraints~\cite{kazdadi2021equality,jalletImplicitDifferentialDynamic2022,howell_altro_2019}. 

\subsection{Adaptive smoothing}
To enforce the convergence towards an optimal (local) solution, it remains crucial to reduce the noise injected via the randomized smoothing across the iterations. 
A first possible strategy \cite{suh2021bundled} consists in relying on Robbins-Monro rule \cite{robbinsmonro1951} by decreasing the variance in a way such that $\sum_k \epsilon_k^2 < \infty$ to guarantee convergence towards a local minimum. 
In this work, we propose to decrease $\epsilon_k$ in a way that adapts to the problem and avoids the smoothing being reduced too quickly, which would lead to performance similar to classical DDP, or too slowly, which would induce an unnecessarily large number of iterations. 
We adapt the smoothing by solving a cascade of randomly smoothed DDP problems.
More concretely, $\| \nabla_u Q \|_{\infty}$ decreases towards 0 when converging towards an optimum.
Whenever $\| \nabla_u Q \|_{\infty}$ is under a given precision threshold $\alpha$ we consider the smooth sub-problem solved and thus reduce the noise intensity $\epsilon$ and the precision threshold $\alpha$, by a factor $\rho$ and $\gamma$ respectively,  before solving the next sub-problem.
Typically, $\rho$ and $\gamma$ are taken equal and set to the value $2$.
This adaptive scheme is summarized in Alg.~\ref{alg:RDDP} and is generic, so it could be transferred to any algorithm using randomized smoothing.

\section{Experiments}\label{sec:exp}

\begin{figure}[t]
    \centering
    \includegraphics[width=0.45 \textwidth]{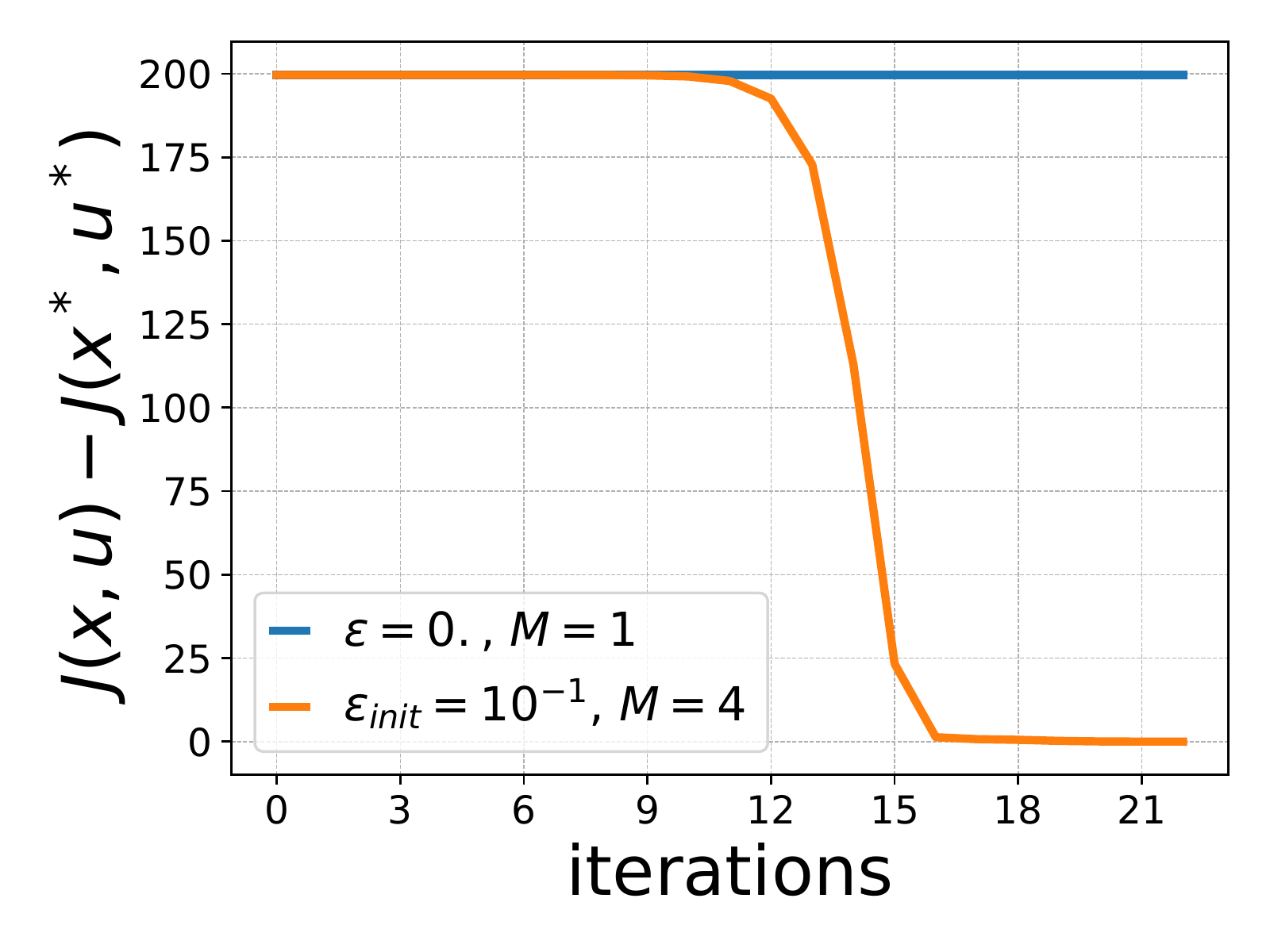}
    \includegraphics[width=0.45 \textwidth]{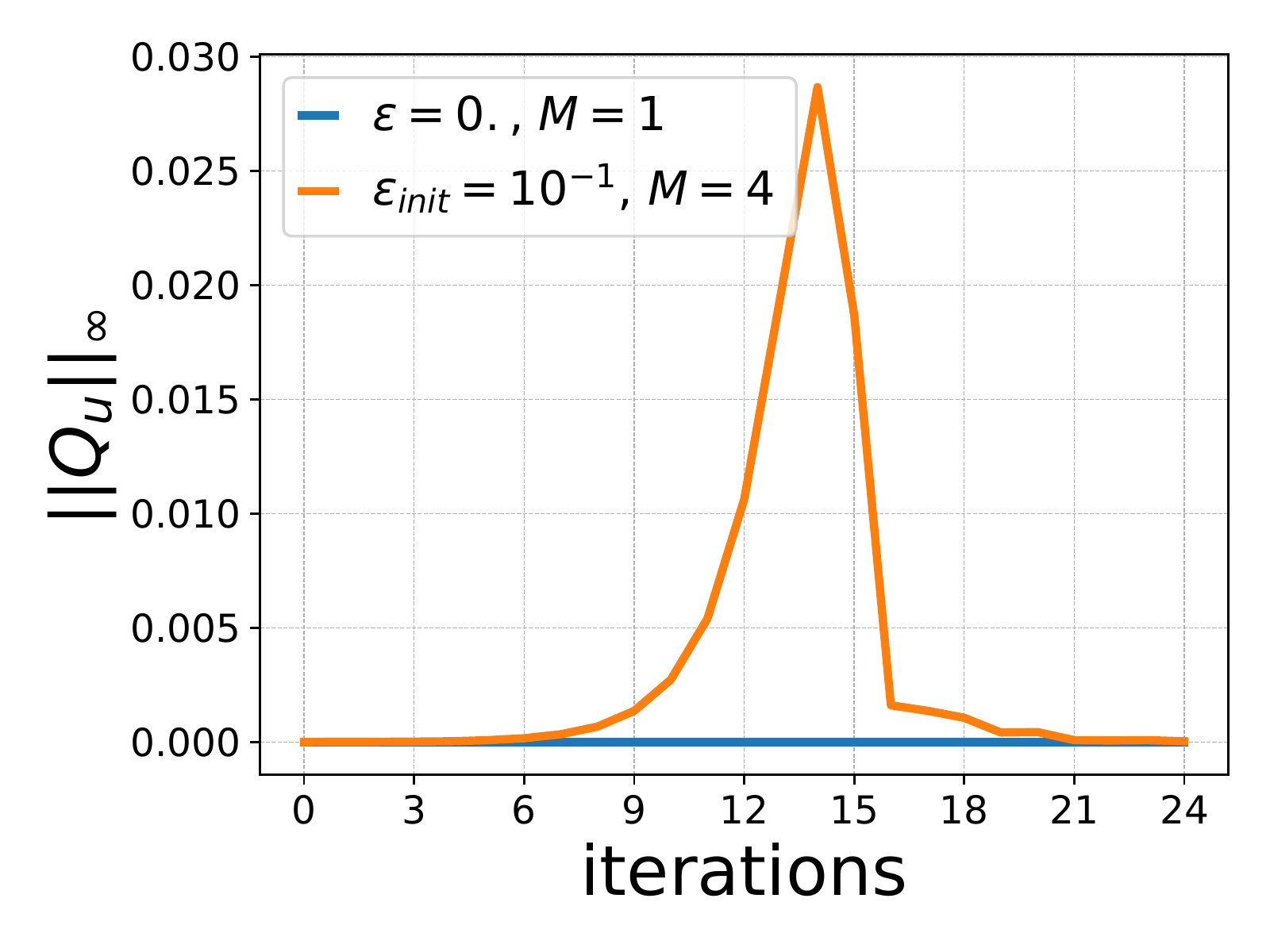}
    \caption{Randomized smoothing allows to escape from the local optima $(x,u)=0$ for the inverted pendulum (see Fig~\ref{fig:Qfunc_pendulum}) and the system reaches the upward position (orange). Without randomized smoothing, the gradients are null and the system remains stuck in the downward position (blue).}
    \label{fig:pendulum_cost}
\end{figure}

In this section, we demonstrate the practical benefits of our randomized DDP algorithms on a set of different systems, ranging from simple toy examples to real robotic systems. We first examine the underlying mechanisms and pendulum and a cube and extend them afterward to a quadrotor and a quadrupedal robot.
Our implementation is based on the open-source frameworks Crocoddyl \cite{mastalli2020crocoddyl} for the DDP algorithm and on Pinocchio~\cite{carpentier-sii19} and \cite{carpentier2018analytical,le2021differentiable} for the derivatives of the dynamics with and without contacts.

\subsection{Avoiding local optima of smooth dynamics}
We consider the task of raising a pendulum from the downward ($\theta = 0$) to the upward vertical position ($\theta = \pi$). The system's state is characterized by the angle $\theta$, representing the deviation of the pendulum bar from the vertical axis, and by the angular velocity $\dot{\theta}$. The tip's position also called the end-effector, can be determined through a trigonometric relation and is denoted as $p(\theta)$, where $p^*$ is the desired position.
We optimize the cost function
\begin{equation}
    J(\boldsymbol{x},\boldsymbol{u}) = w_p \| p(\theta_N) - p^* \|^2 +  \sum_{t=0}^N w_u \| u_t - u^* \|^2, \label{eq:cost_pendulum}
\end{equation}
over $N=400$ time steps and each with a duration $dt=5 \times 10^{-3}\,$s. The weights are defined as $w_p = 2$, $w_u = 2 \times 10^{-5}$ and the system is described using the generalized coordinates $x=(\theta, \dot{\theta})$ of position and velocity. 
As explained previously, due to the robot kinematics, the term of $J$ involving the distance on the end-effector $ \| p(\theta_N) - p^* \|$ is non-convex leading to multiple potential local optima of the control problem.
We run 24 iterations of the classical DDP and the RDDP algorithm optimizing for a trajectory with 400 time steps to swing the pendulum up.
For the problem \eqref{eq:cost_pendulum}, the solution \mbox{$(u,x)= (0, 0)$} is a local extremum, and the classical DDP algorithm gets stuck at this point (Fig.~\ref{fig:pendulum_cost}). As discussed in Sec.~\ref{sec:rs_rl}, randomized smoothing using a smaller number of only $M=4$ samples per control input to smooth and average the resulting state allows RDDP to avoid this local optimum (Fig.~\ref{fig:Qfunc_pendulum},\ref{fig:pendulum_cost}).
Here, two distinct effects are at work: i) the Randomized Smoothing can smoothen out some local optima, and ii) the noise from the Monte-Carlo estimator helps to escape from unstable critical points as detailed in~\cite{ge2015escaping}. In addition to converging towards better optima in the case of smooth dynamics, we primarily designed RDDP to provide a solution when it comes to the aforementioned issues from non-smooth dynamics (see Sec.~\ref{sec:local_OC} and Appendix~\ref{sec:app-ncp}), shown in the next section. 

\subsection{Controlling non-smooth dynamics}


\begin{figure*}[t]
    \centering
    \includegraphics[width=0.30 \textwidth]{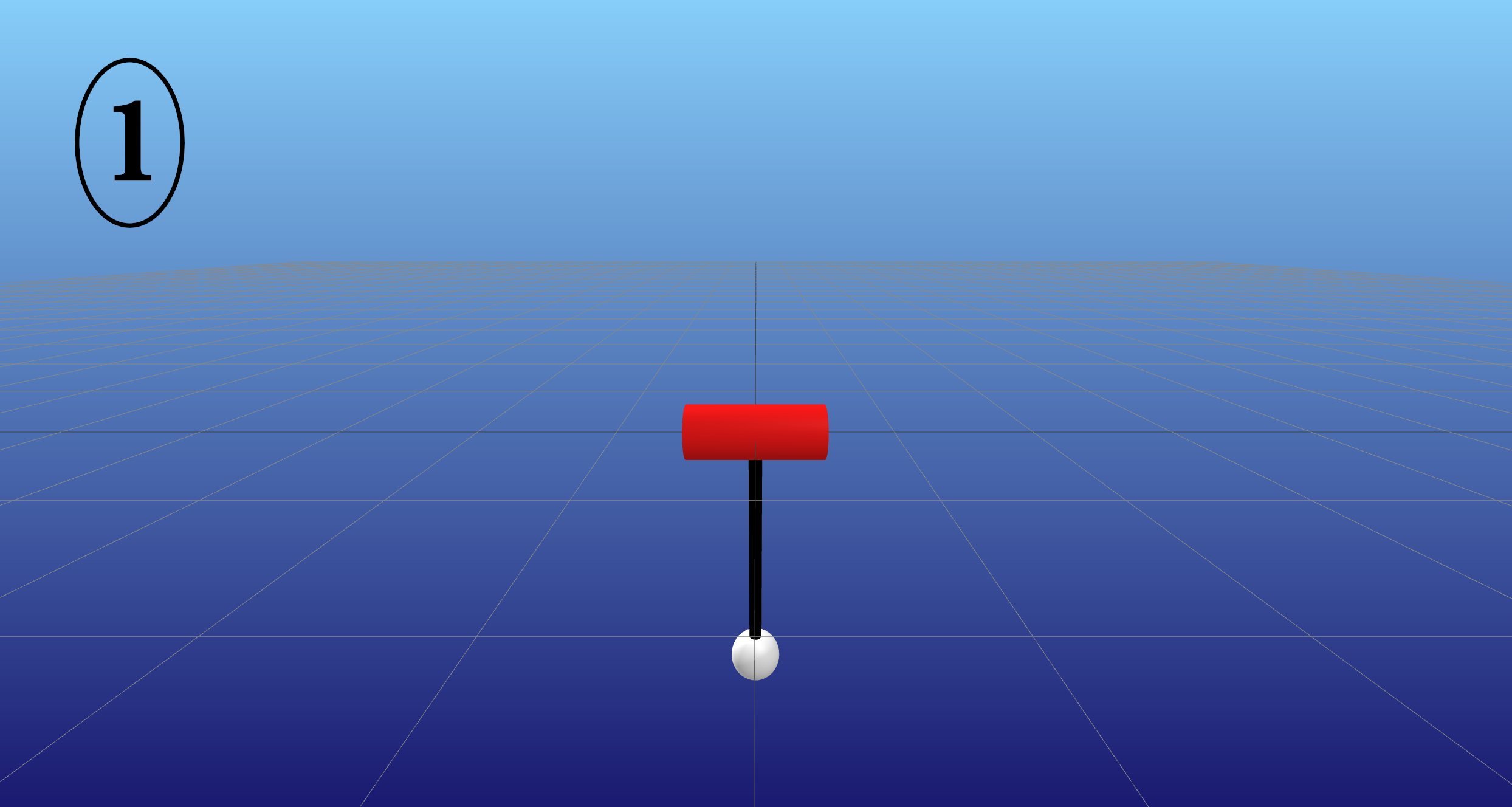}
    \includegraphics[width=0.30 \textwidth]{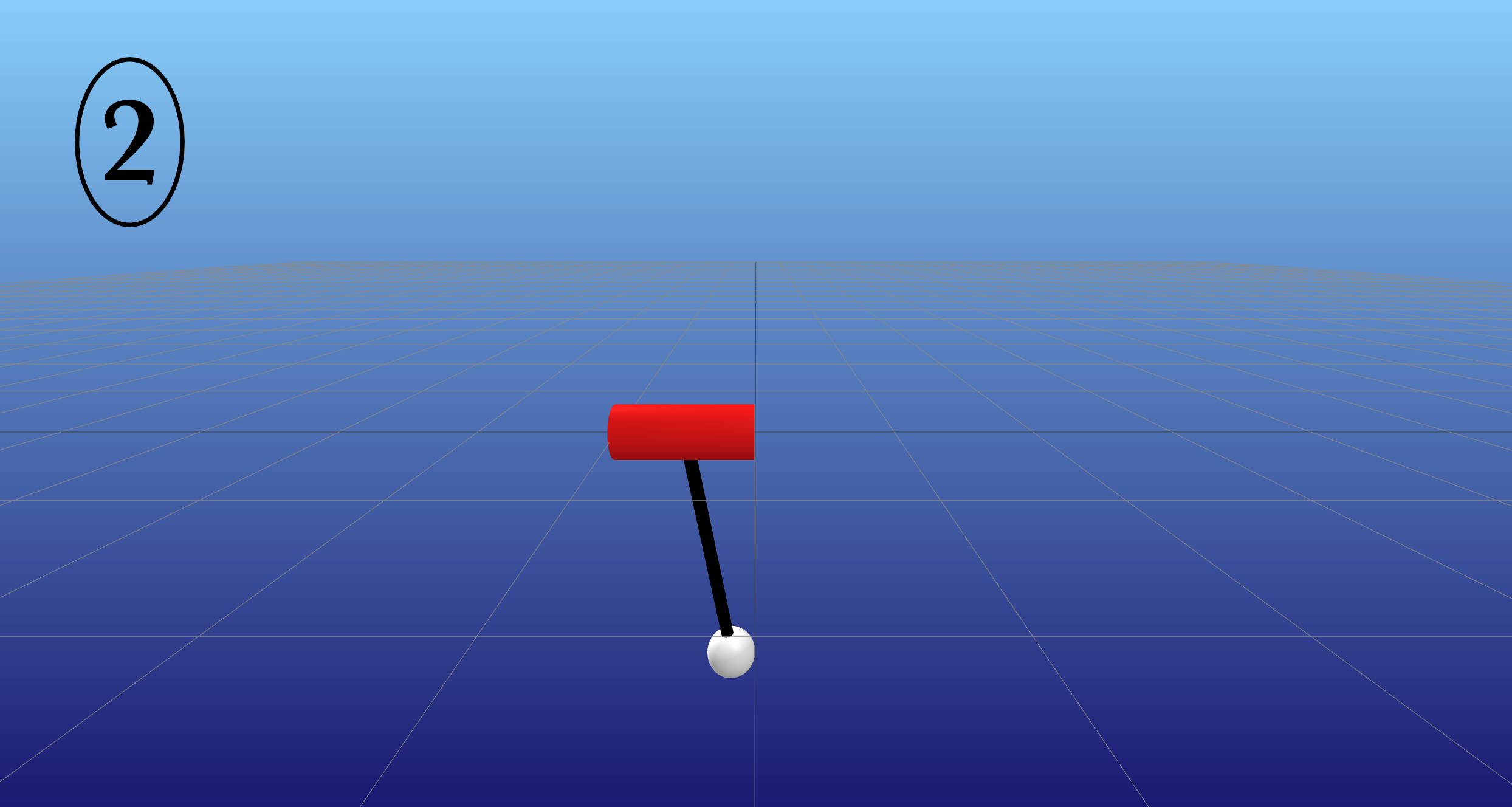}
    \includegraphics[width=0.30 \textwidth]{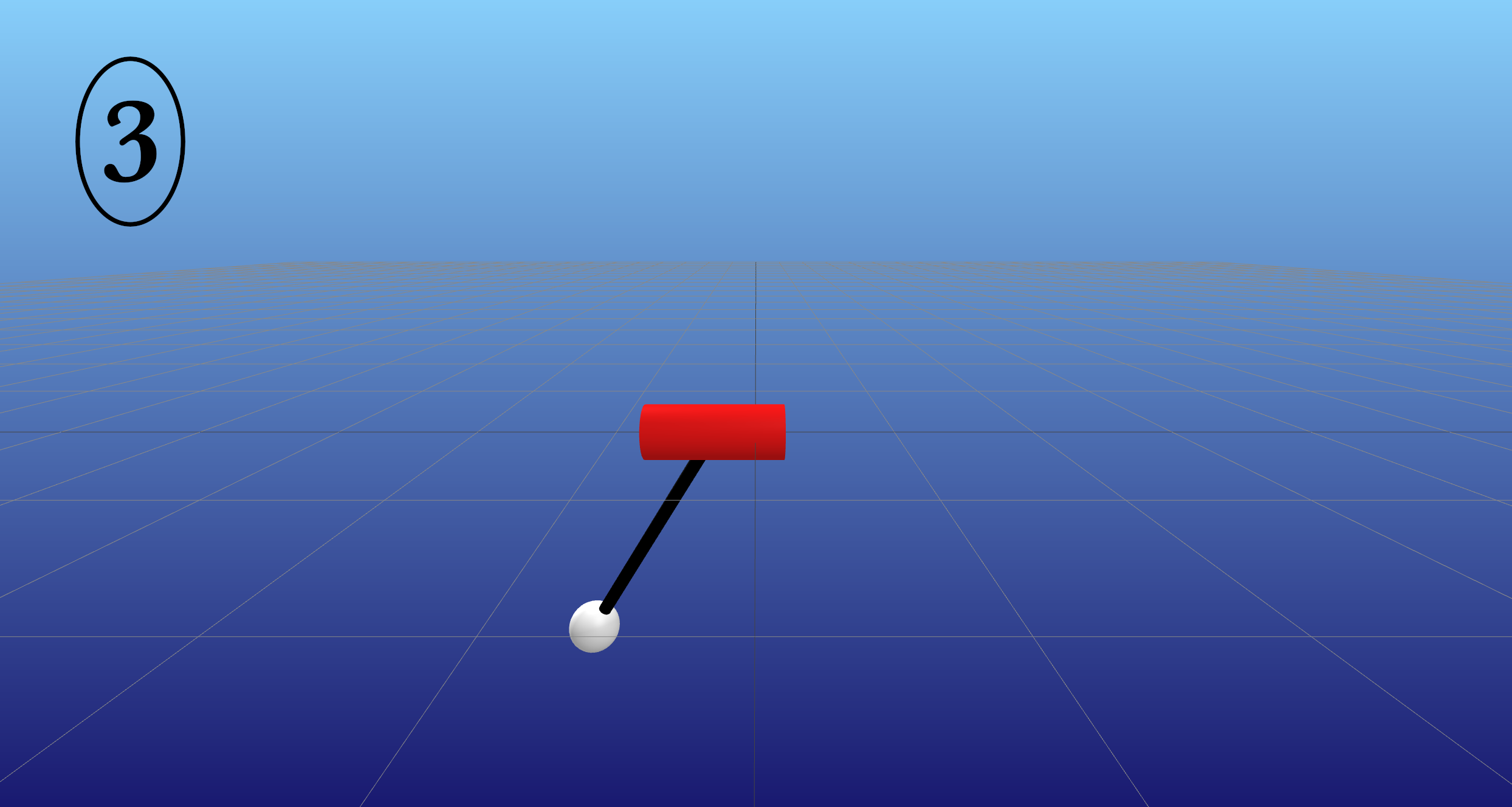} \\
    \includegraphics[width=0.30 \textwidth]{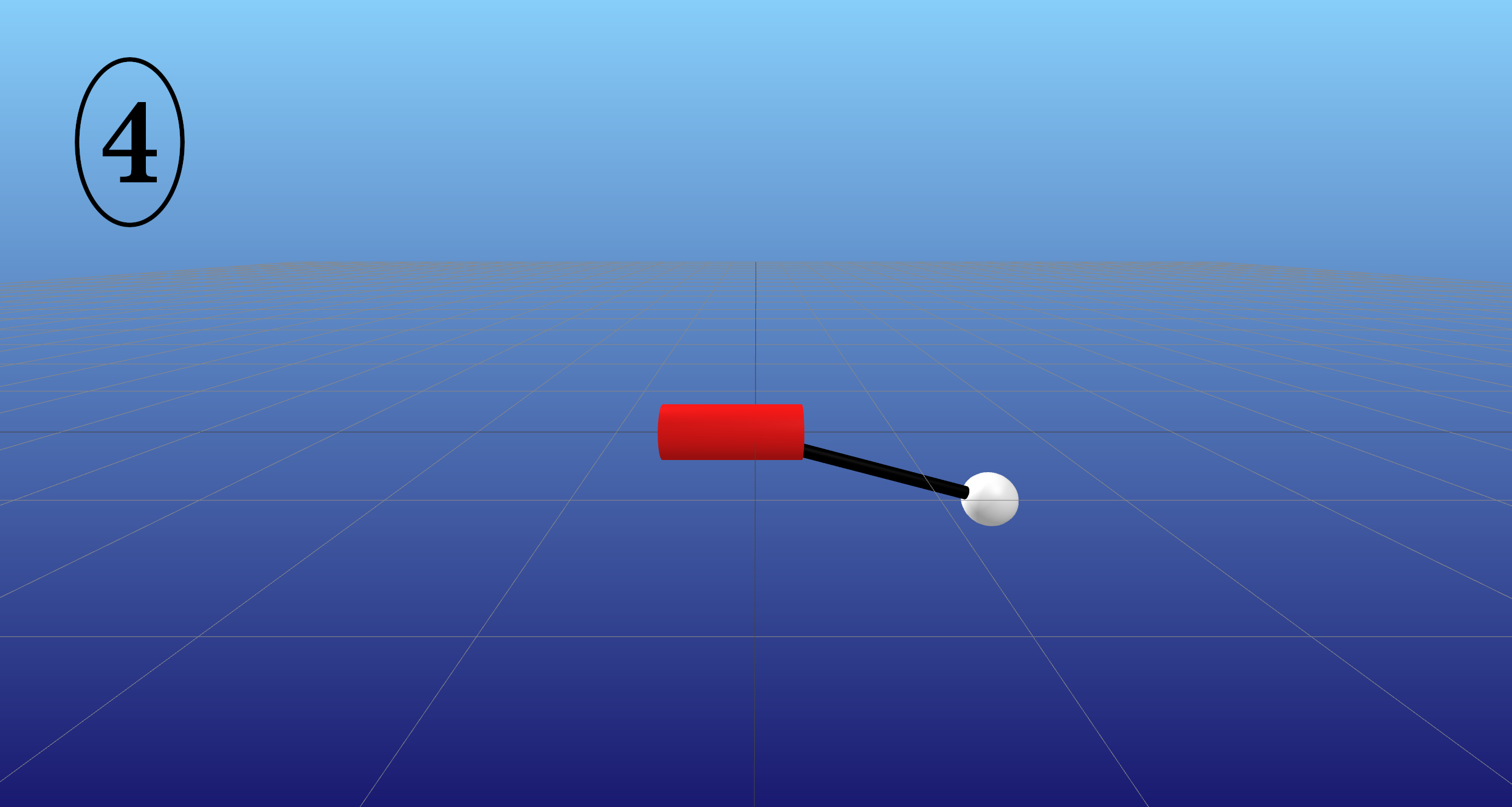}
    \includegraphics[width=0.30 \textwidth]{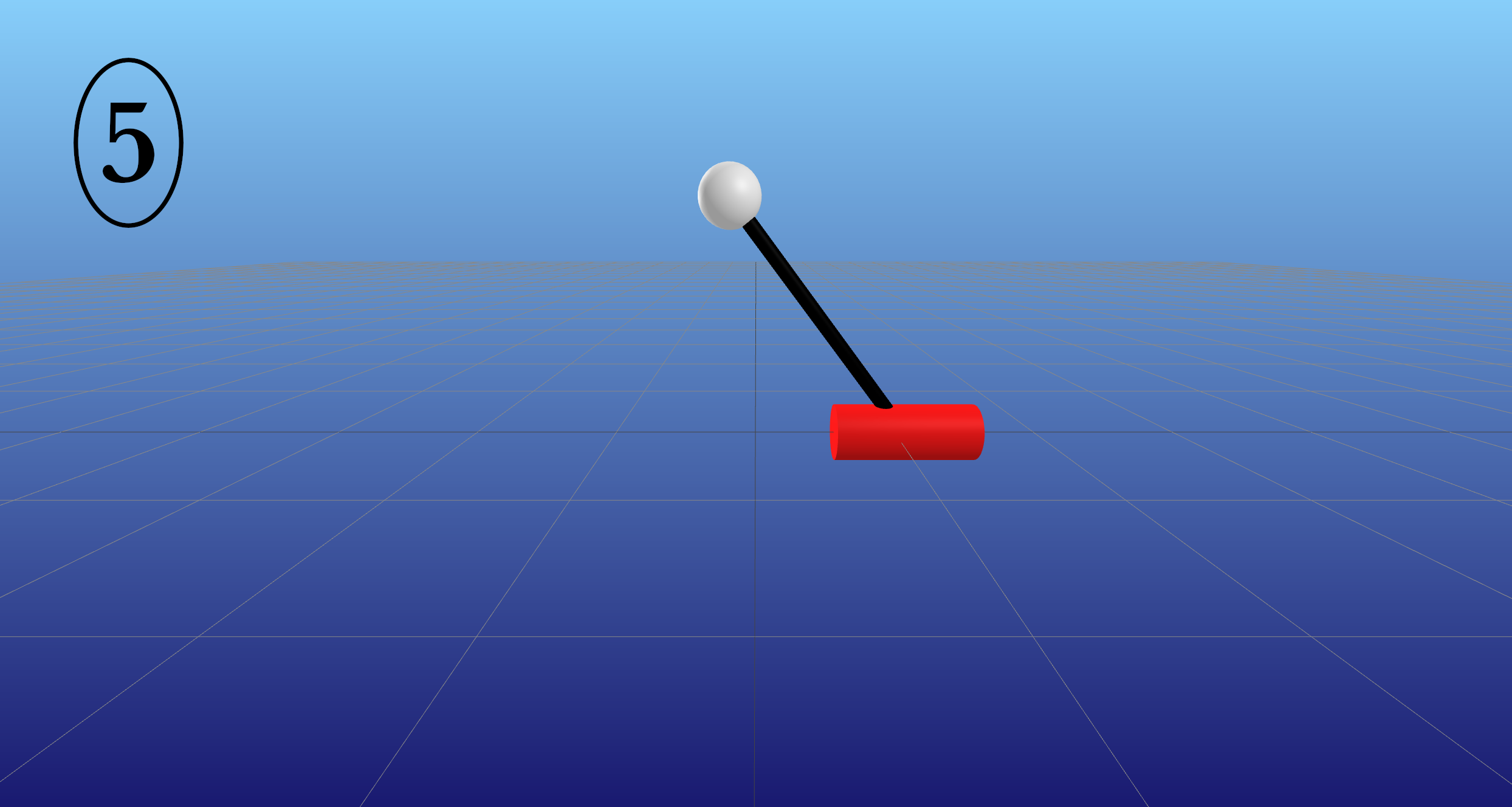}
    \includegraphics[width=0.30 \textwidth]{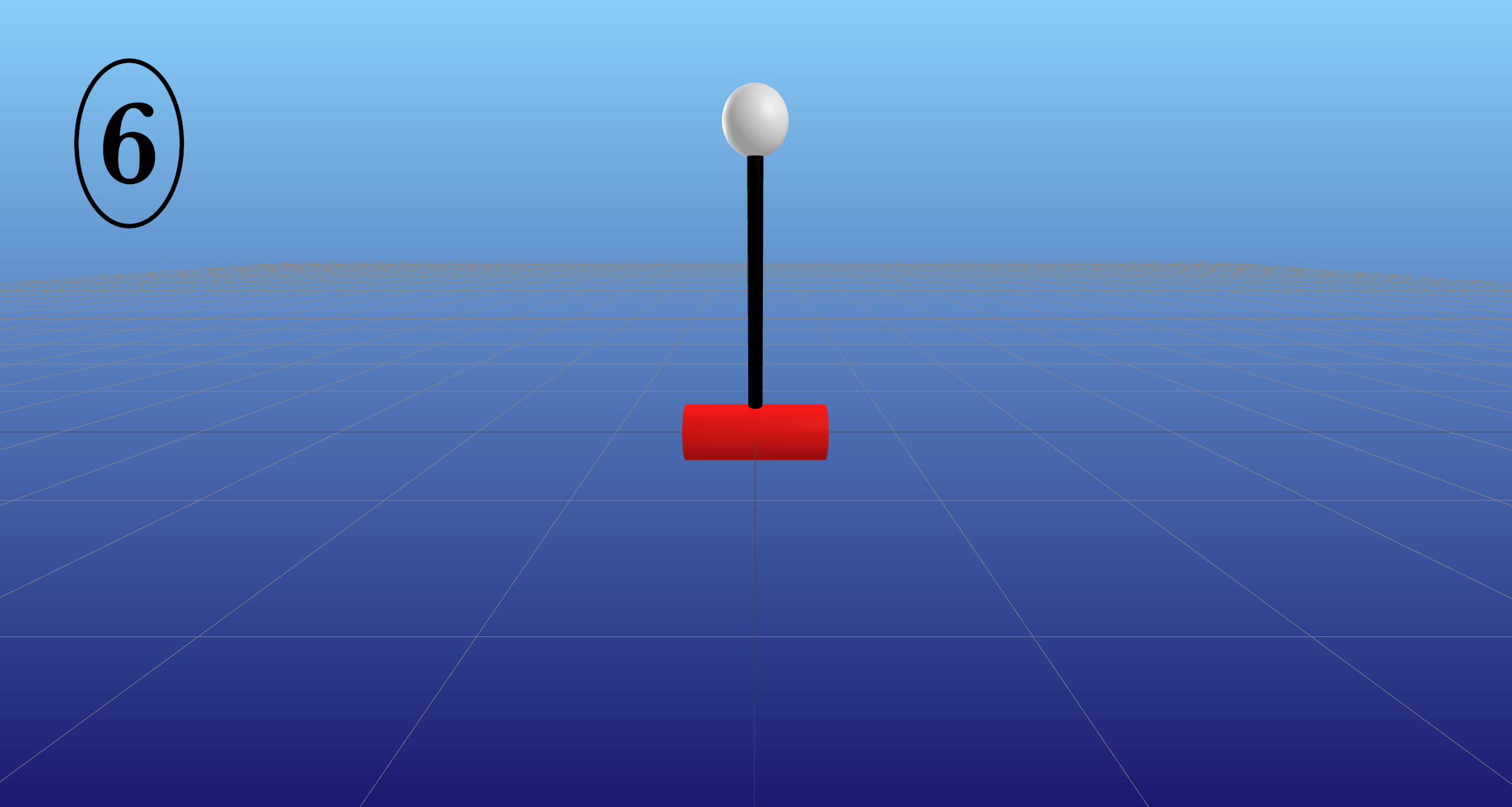} \\
    \vspace{0.5cm}
    \includegraphics[width=0.48 \linewidth]{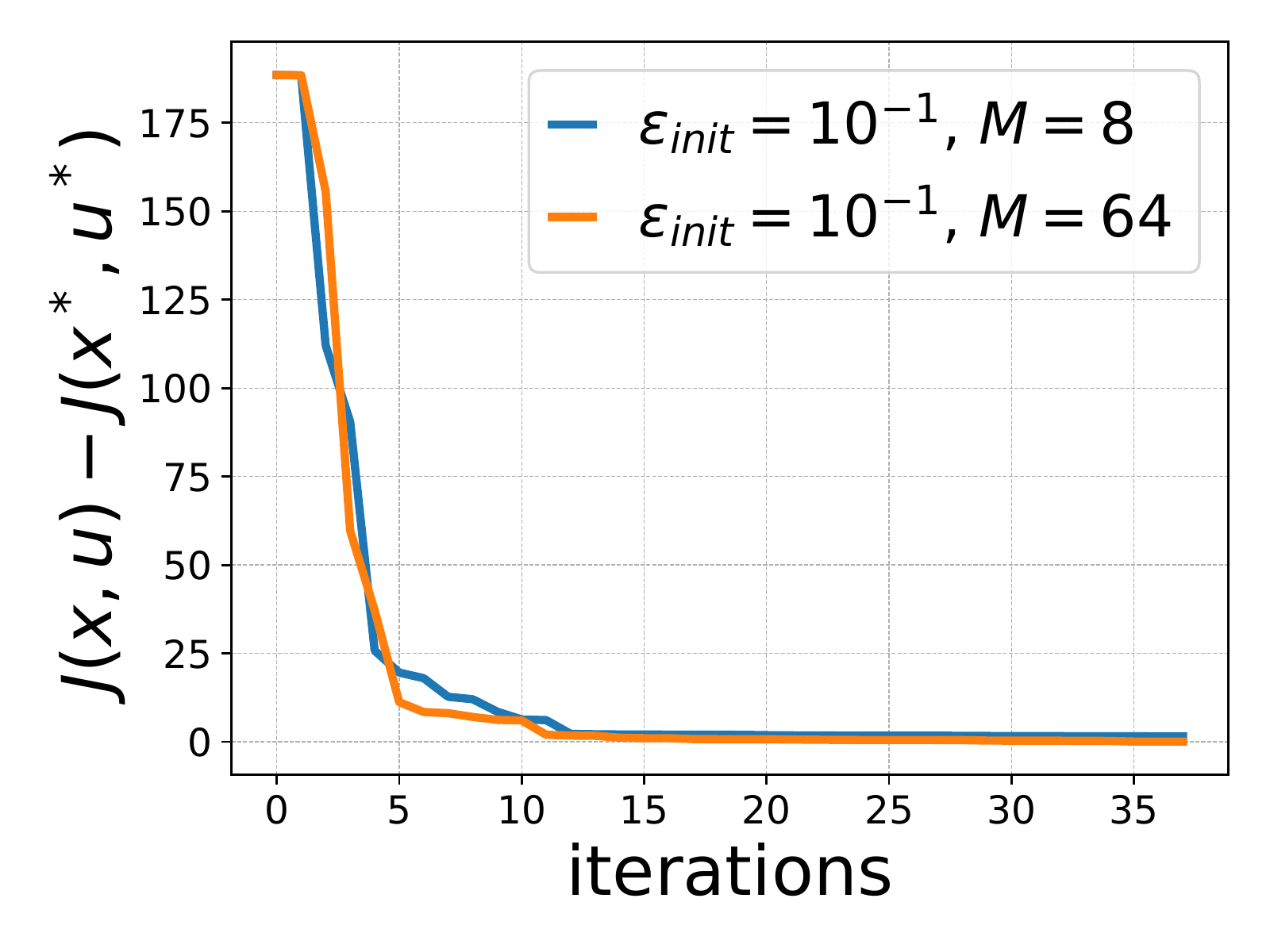}
    \includegraphics[width=0.48 \linewidth]{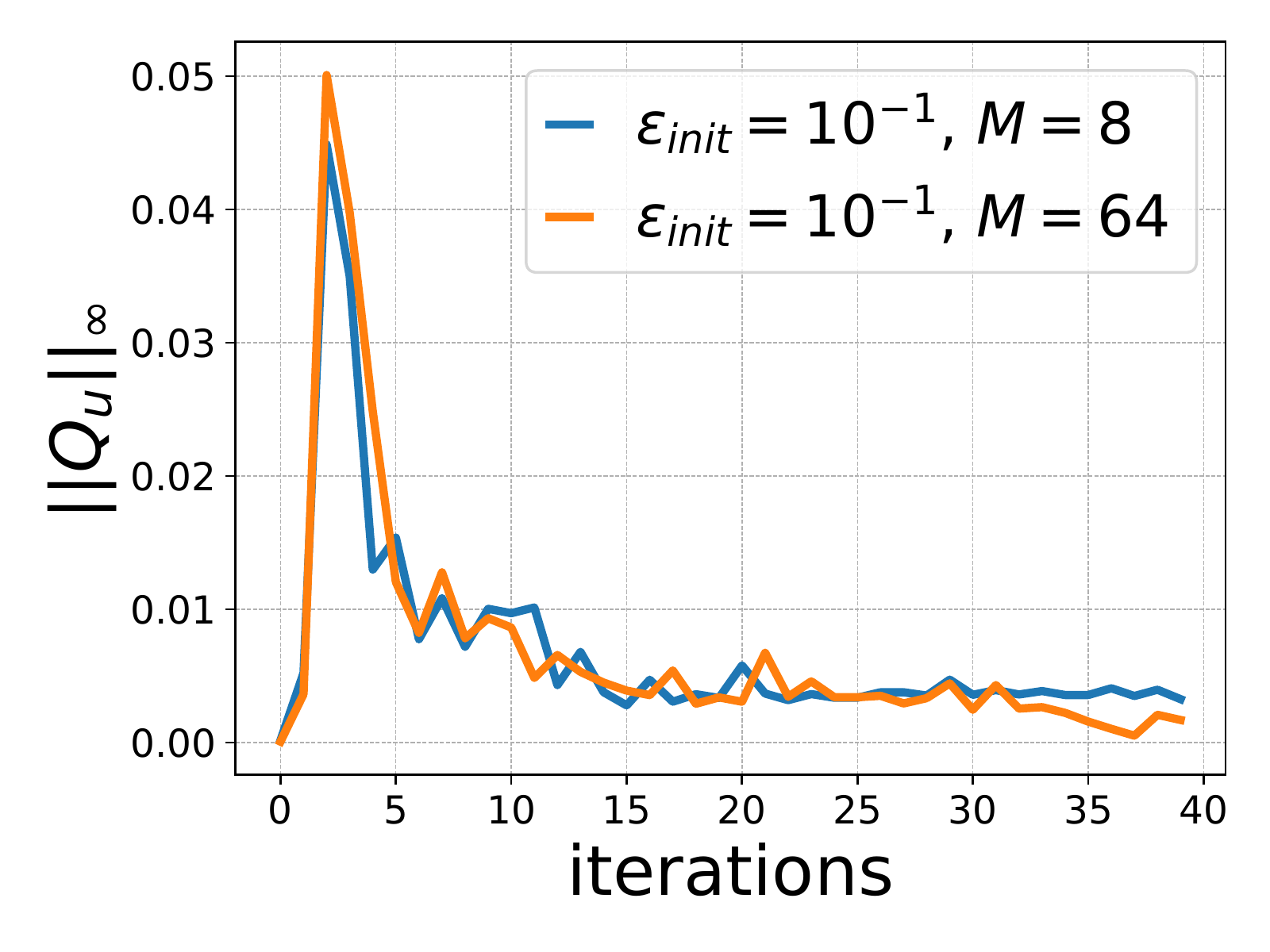}
\caption{\textbf{Top:} Application of RDDP to precisely control a cartpole movement requiring overcoming dry friction on the joints and moving the pole straight upward.
    \textbf{Bottom:} Using RDDP for the cartpole task, we compare the influence of the sample size on the solution. Only a few samples (here M=8) are necessary to get reasonable results. Increasing the number of samples in the first order Monte-Carlo estimator of the gradients leads to very marginal improvement.}
    \label{fig:cartpole}
    \vspace{-0.0cm}
\end{figure*}

\begin{figure}[t]
    \centering
    \includegraphics[width=0.30 \linewidth]{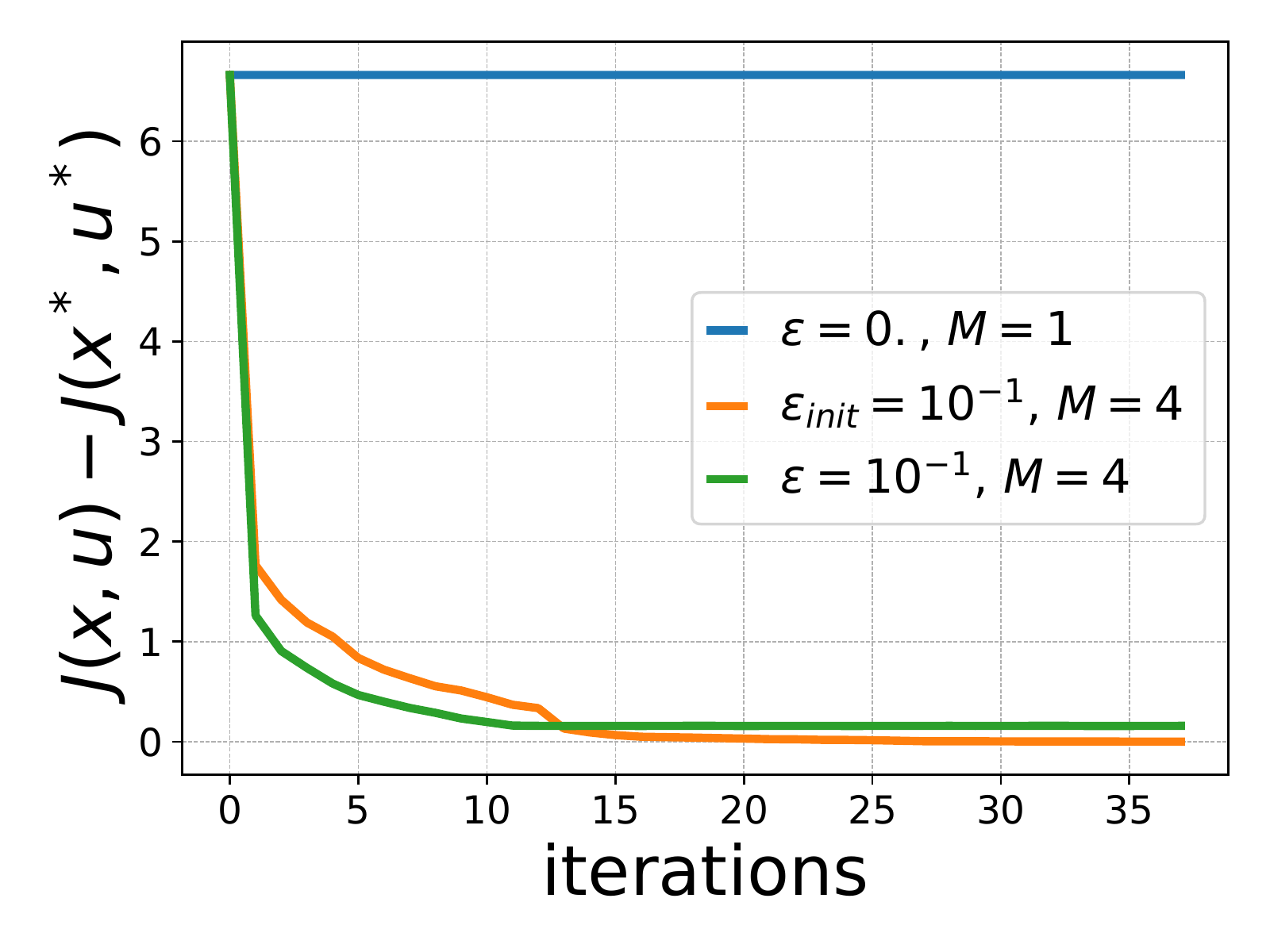}
    \includegraphics[width=0.30\linewidth]{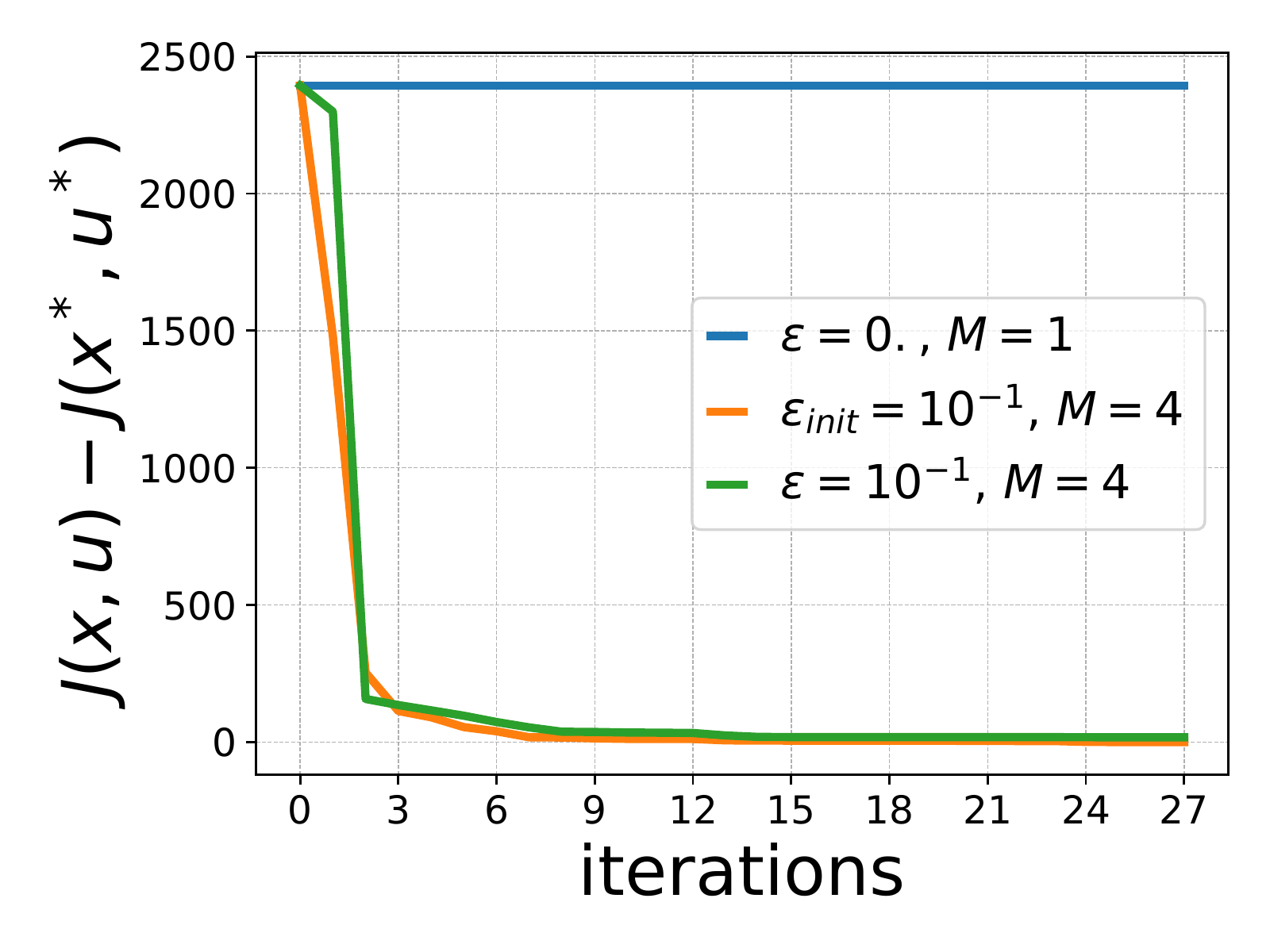}
    \includegraphics[width=0.30 \linewidth]{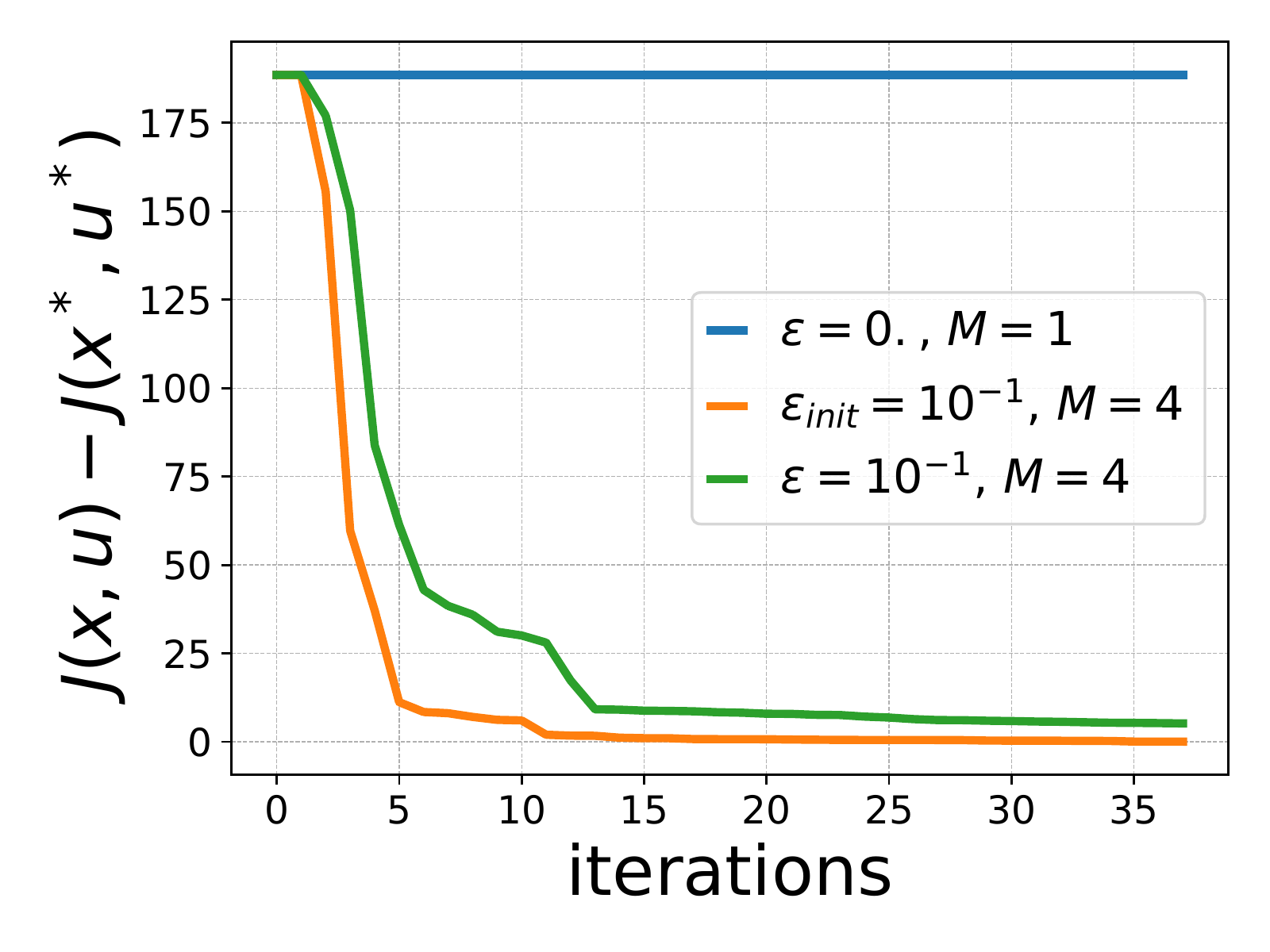}
    \includegraphics[width=0.30 \linewidth]{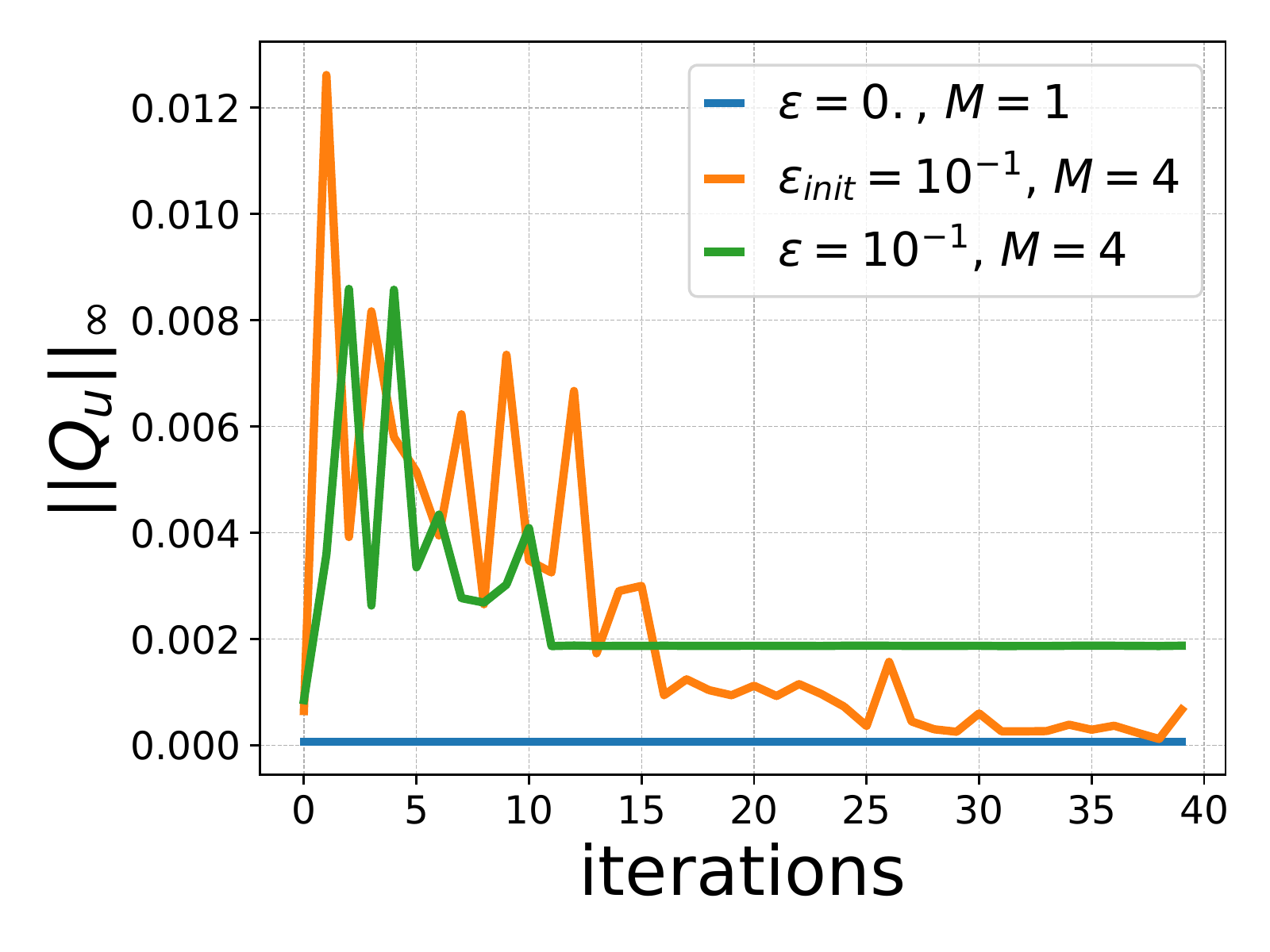}
    \includegraphics[width=0.30\linewidth]{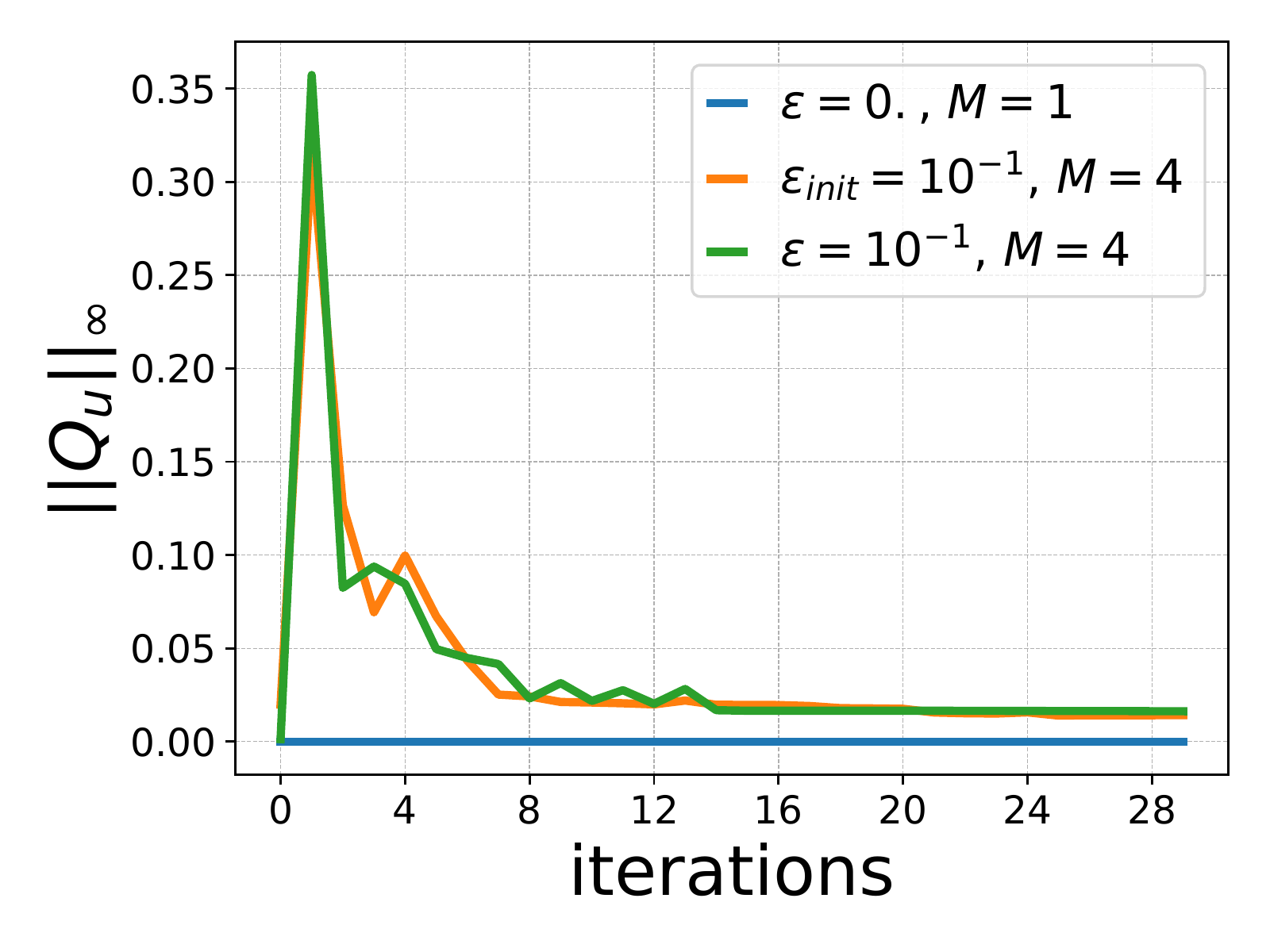}
    \includegraphics[width=0.30 \linewidth]{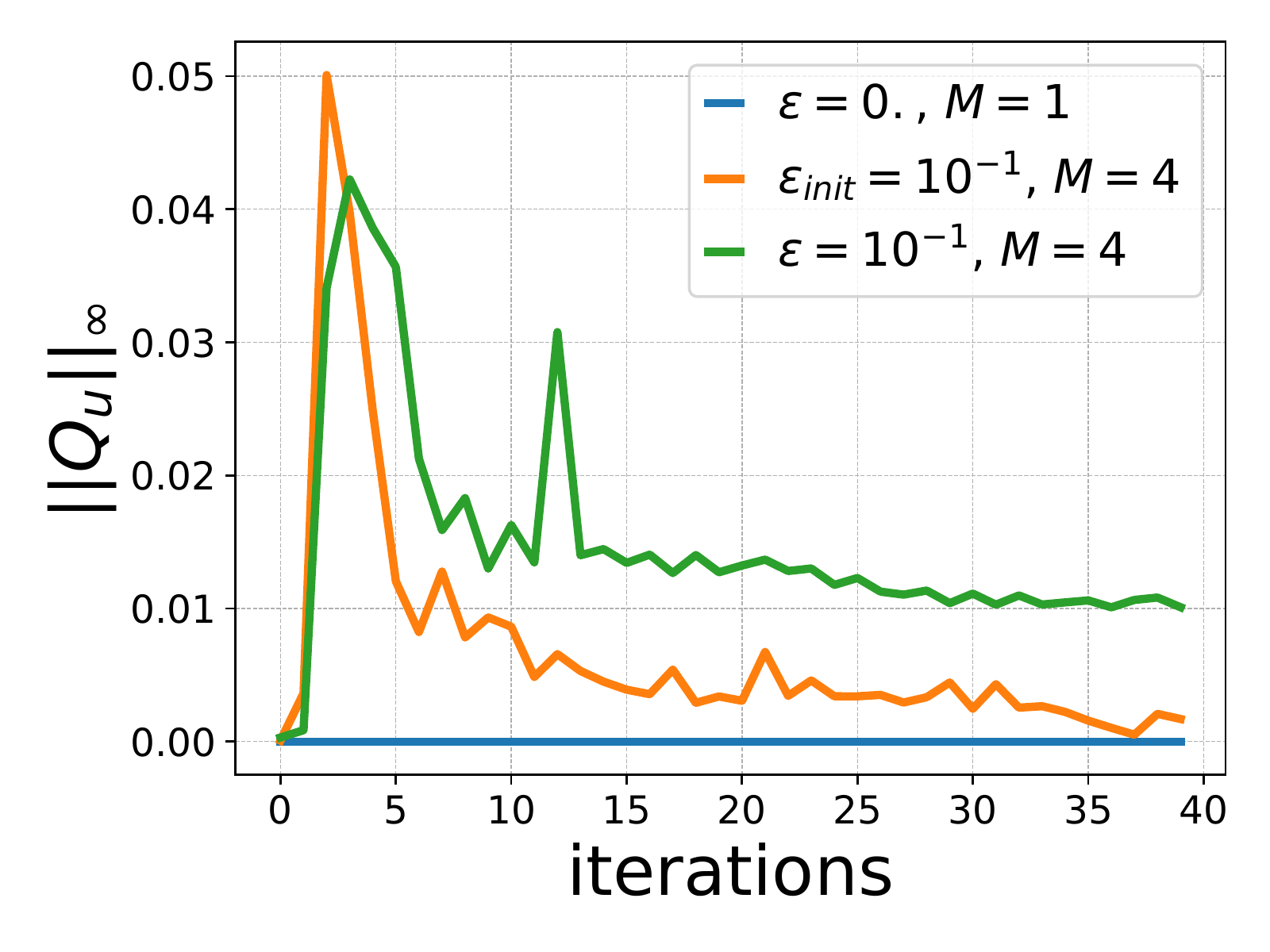}
    \caption{This figure shows the effect of our adaptive noise scheduling (orange) RDDP in comparison against fixed noise (green) RDDP and classical DDP with no noise (blue) for three different tasks.
    \textbf{Left:} RDDP solves tasks requiring breaking contacts such as the cube-lifting from Fig.~\ref{fig:cube_schema}  \textbf{Middle:} Similarly, the approach solves classical control problems on underactuated systems such as an inverted double pendulum or, \textbf{Right:} a cartpole with dry frictions on the joints. While both RDDP variants solve the task contrary to classical DDP, observe the improvement on the convergence criteria $\|Q_u \|_{\infty}$ for the adaptive noise scheduling (orange).}
    \label{fig:cube_double_cartpole}
\end{figure}

\begin{figure*}[h!]
    \centering
    \includegraphics[width=0.30 \textwidth]{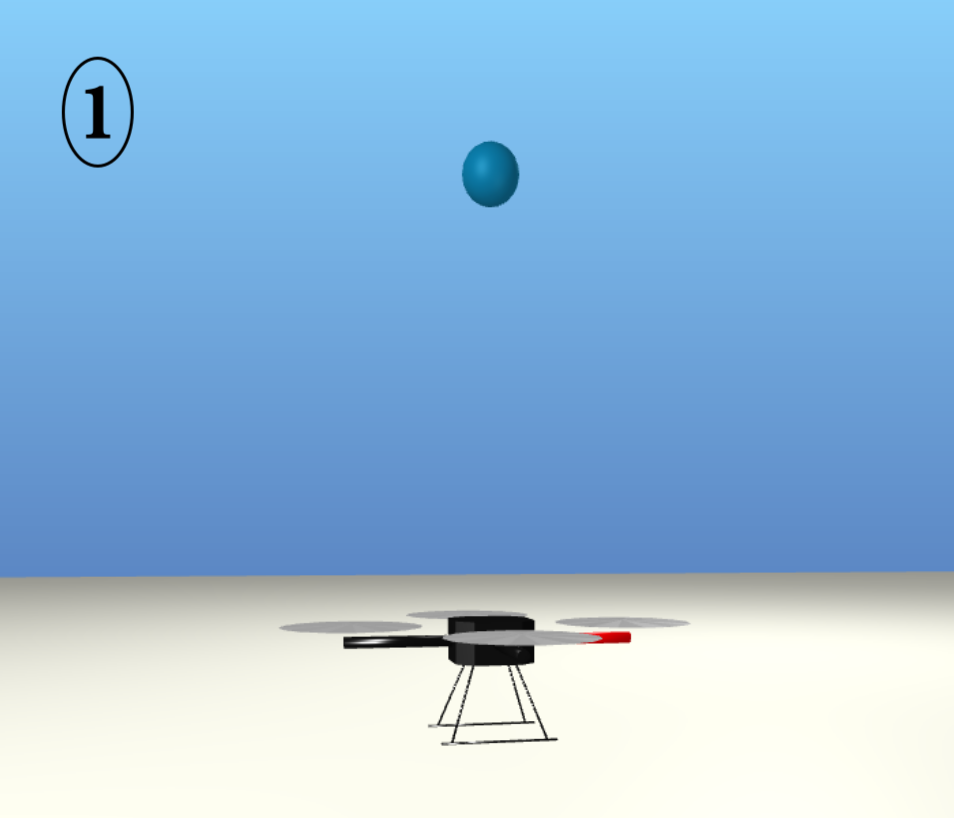}
    \includegraphics[width=0.30 \textwidth]{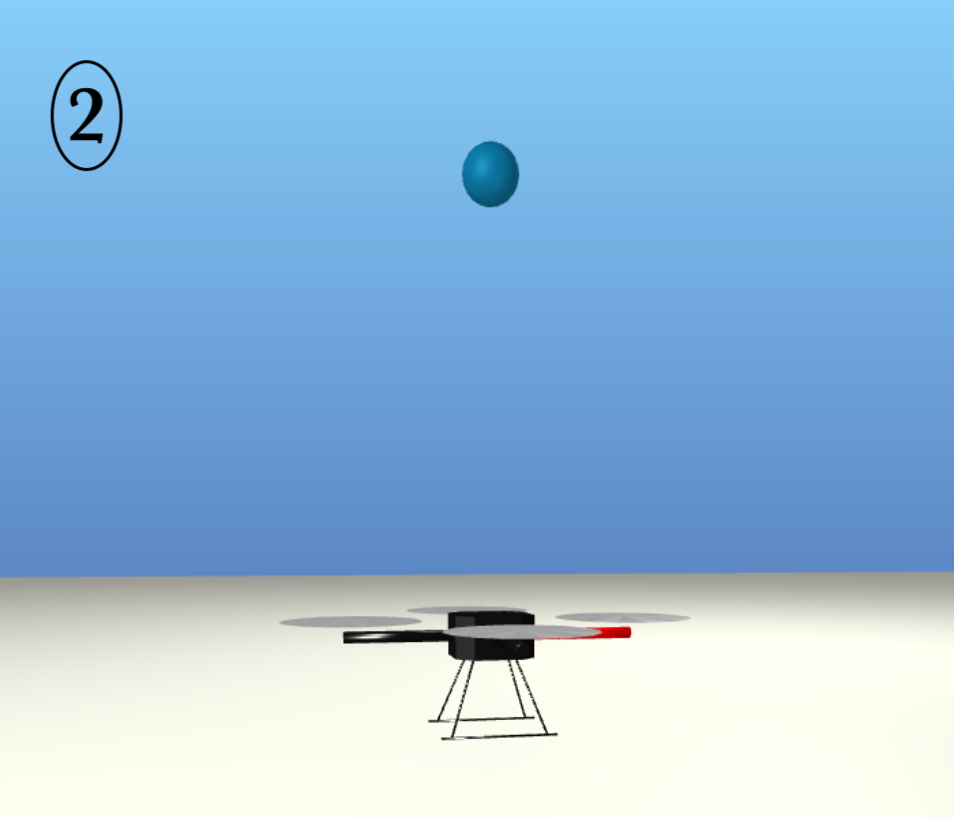}
    \includegraphics[width=0.30 \textwidth]{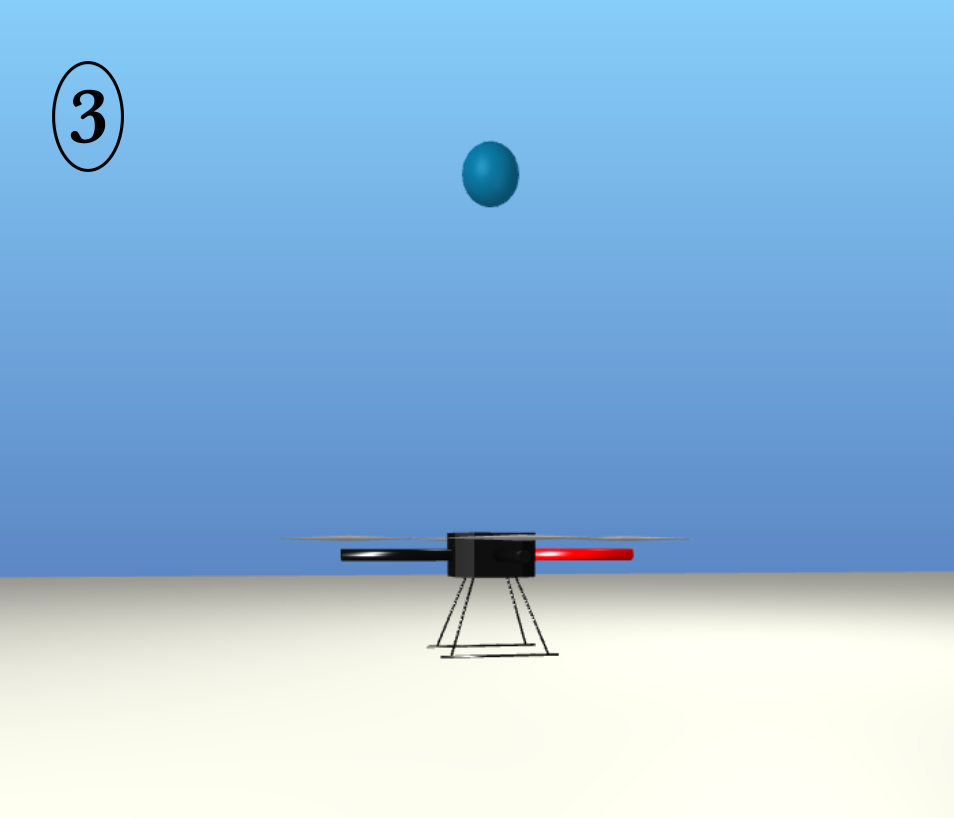} \\
    \includegraphics[width=0.30 \textwidth]{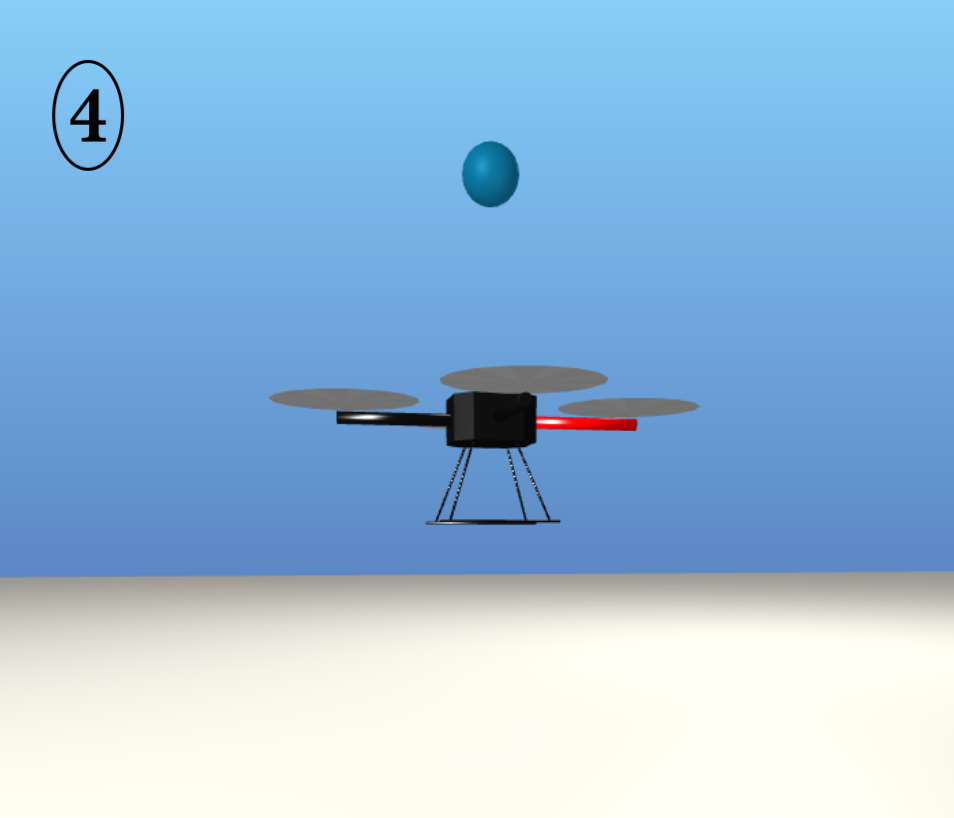}
    \includegraphics[width=0.30 \textwidth]{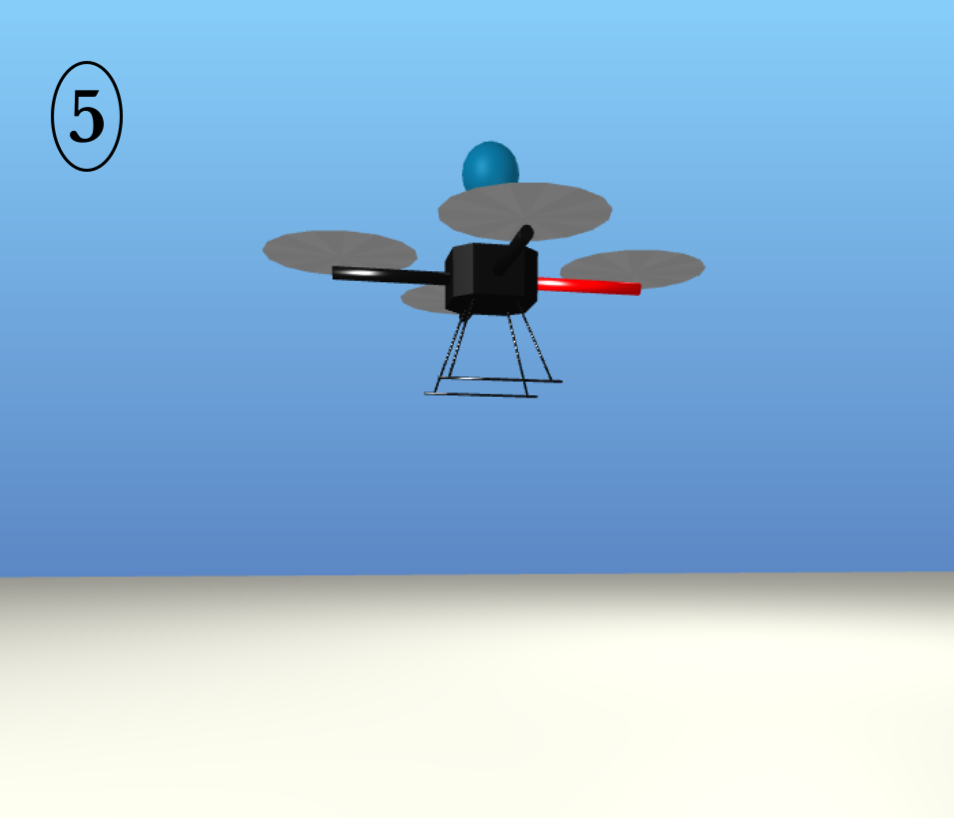}
    \includegraphics[width=0.30 \textwidth]{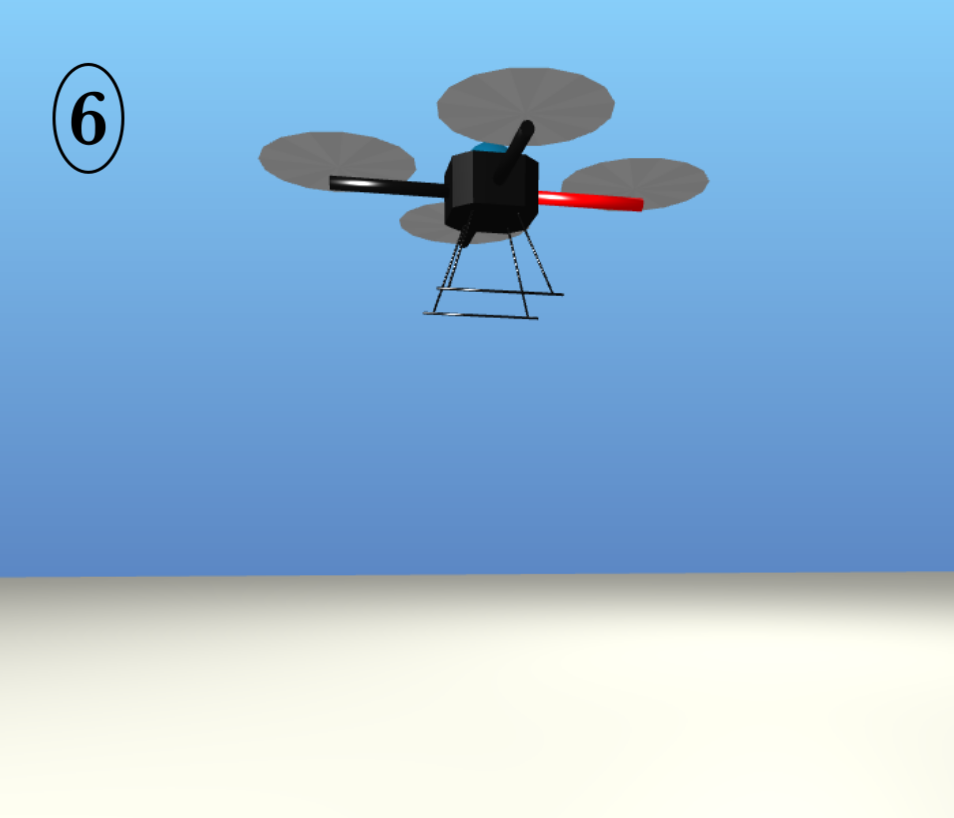} \\
    \vspace{0.5cm}
    \includegraphics[width=0.45 \linewidth]{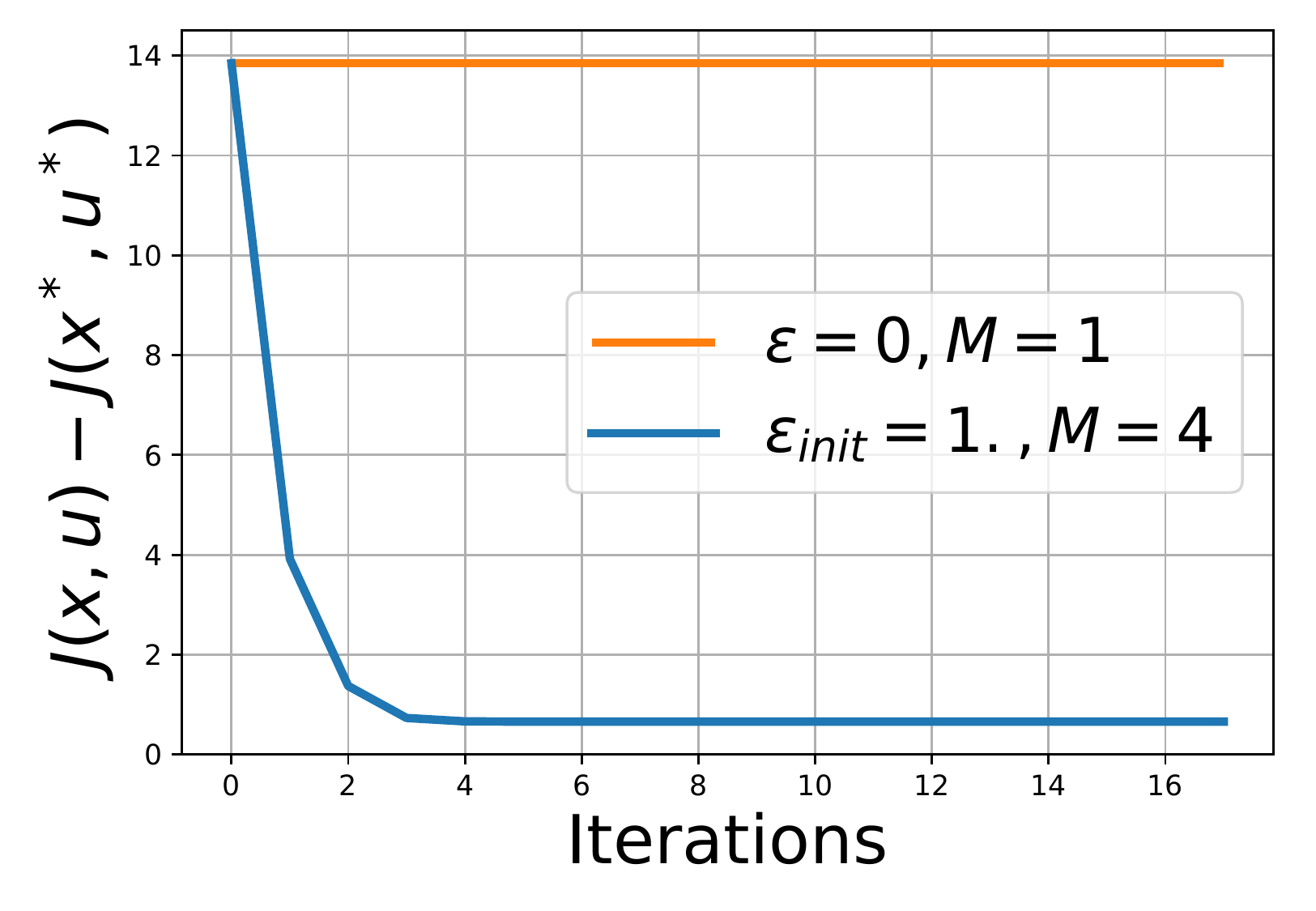}
    \includegraphics[width=0.45 \linewidth]{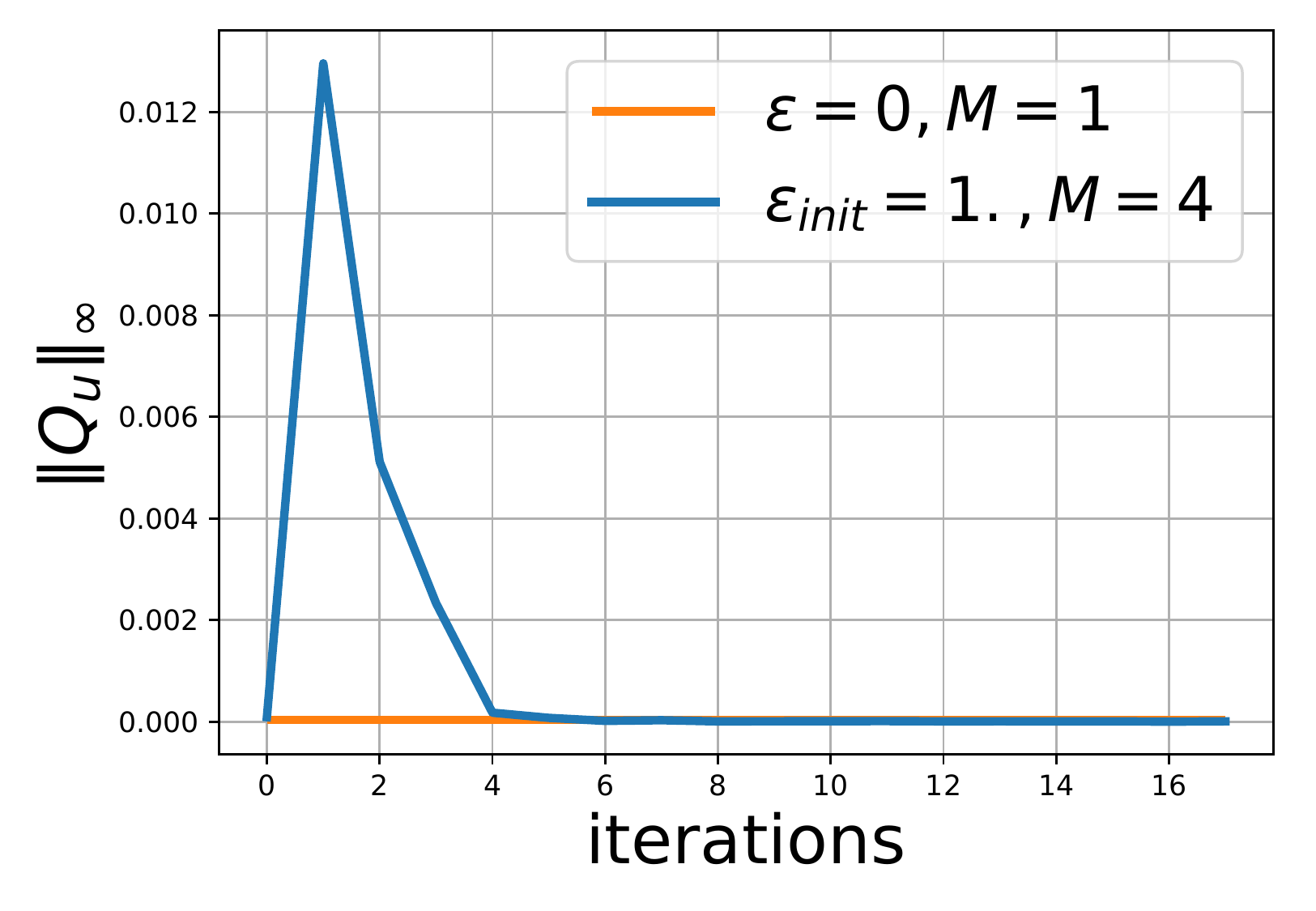}
    \caption{Comparison of RDDP (blue) against classical DDP (orange) to plan the take-off of a drone.
    \textbf{Top:} Sequence for take-off achieved by RDDP which allows to reach the goal position marked with a blue sphere on the pictures.
    \textbf{Bottom:} On the opposite, unilateral contacts induce null gradients which lead to the failure of classical DDP.}
    \label{fig:pictures_quad}
    \vspace{-0.0cm}
\end{figure*}

\begin{figure*}[h!]
    \centering
    \includegraphics[width=0.30 \textwidth]{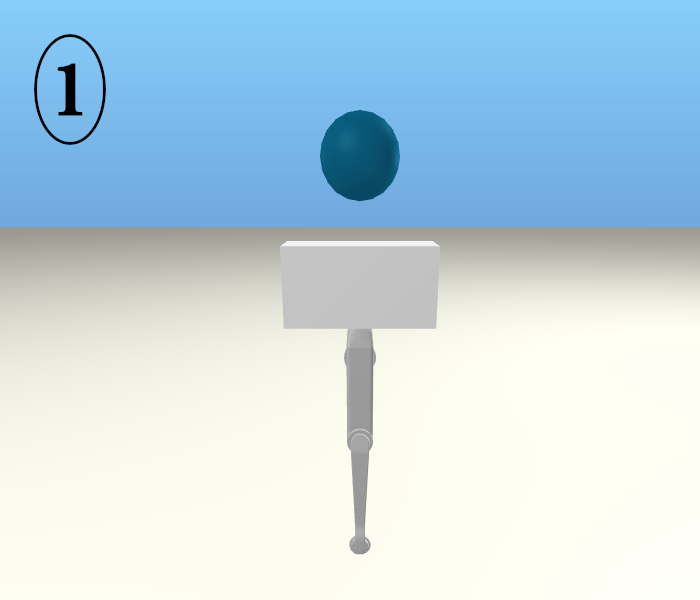}
    \includegraphics[width=0.30 \textwidth]{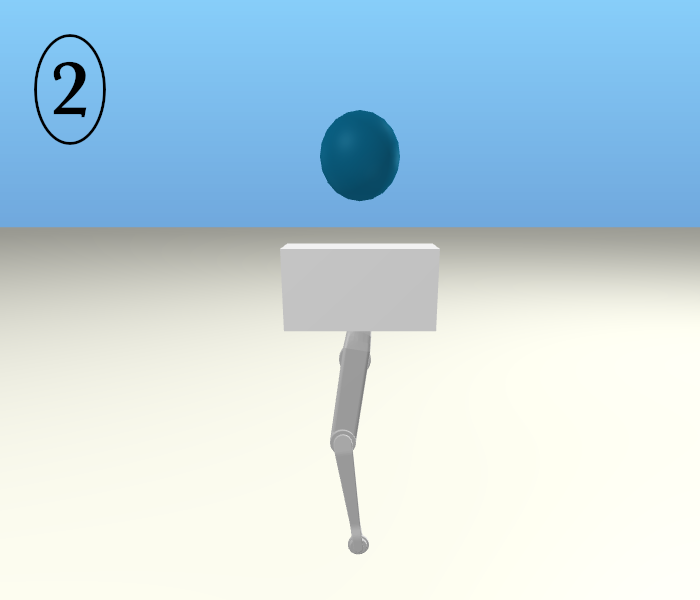}
    \includegraphics[width=0.30 \textwidth]{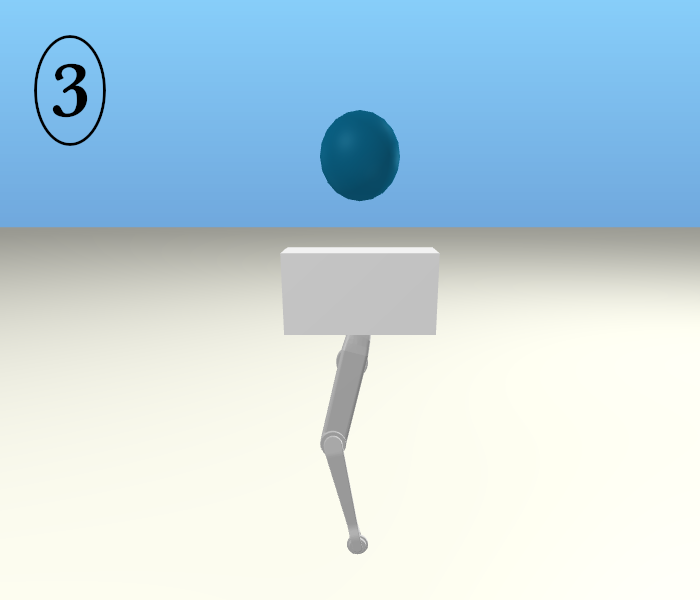} \\
    \includegraphics[width=0.30 \textwidth]{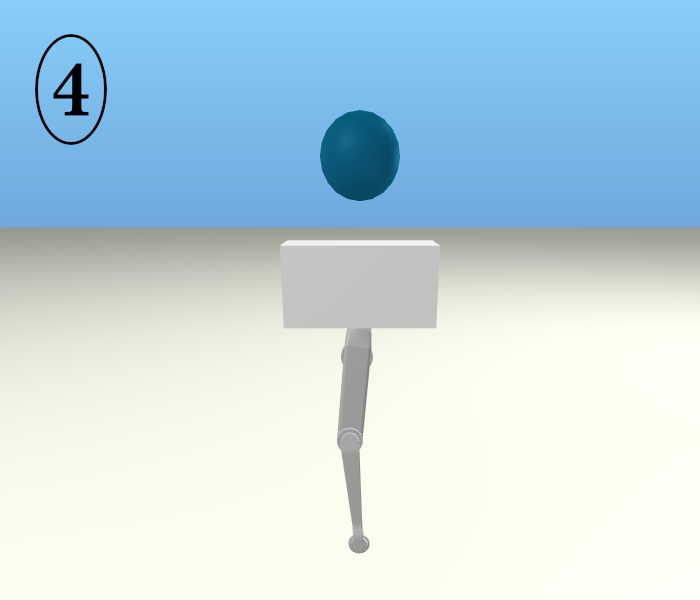}
    \includegraphics[width=0.30 \textwidth]{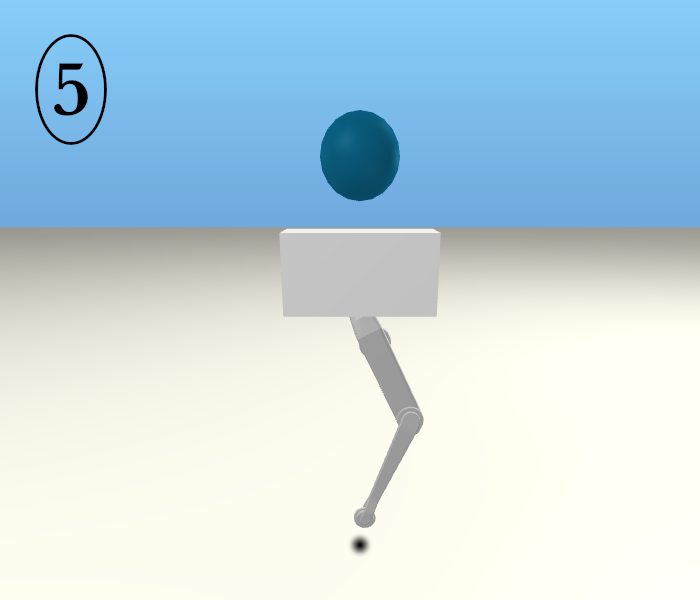}
    \includegraphics[width=0.30 \textwidth]{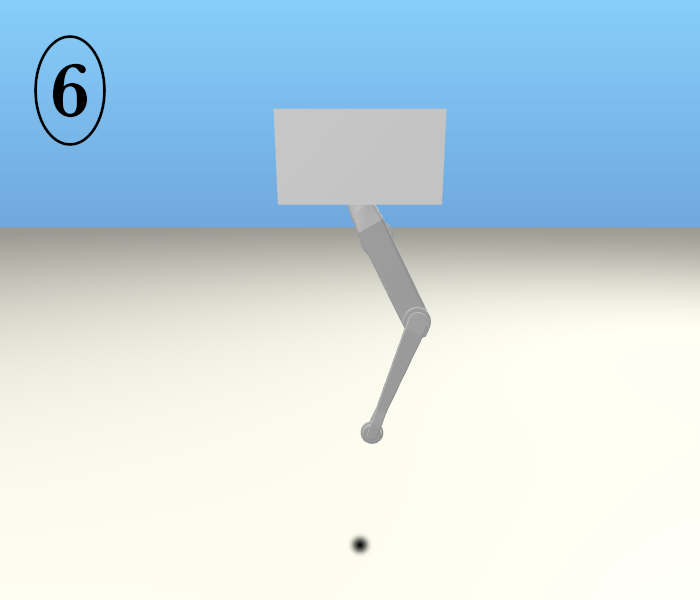} \\
    \vspace{0.5cm}
    \includegraphics[width=0.45 \linewidth]{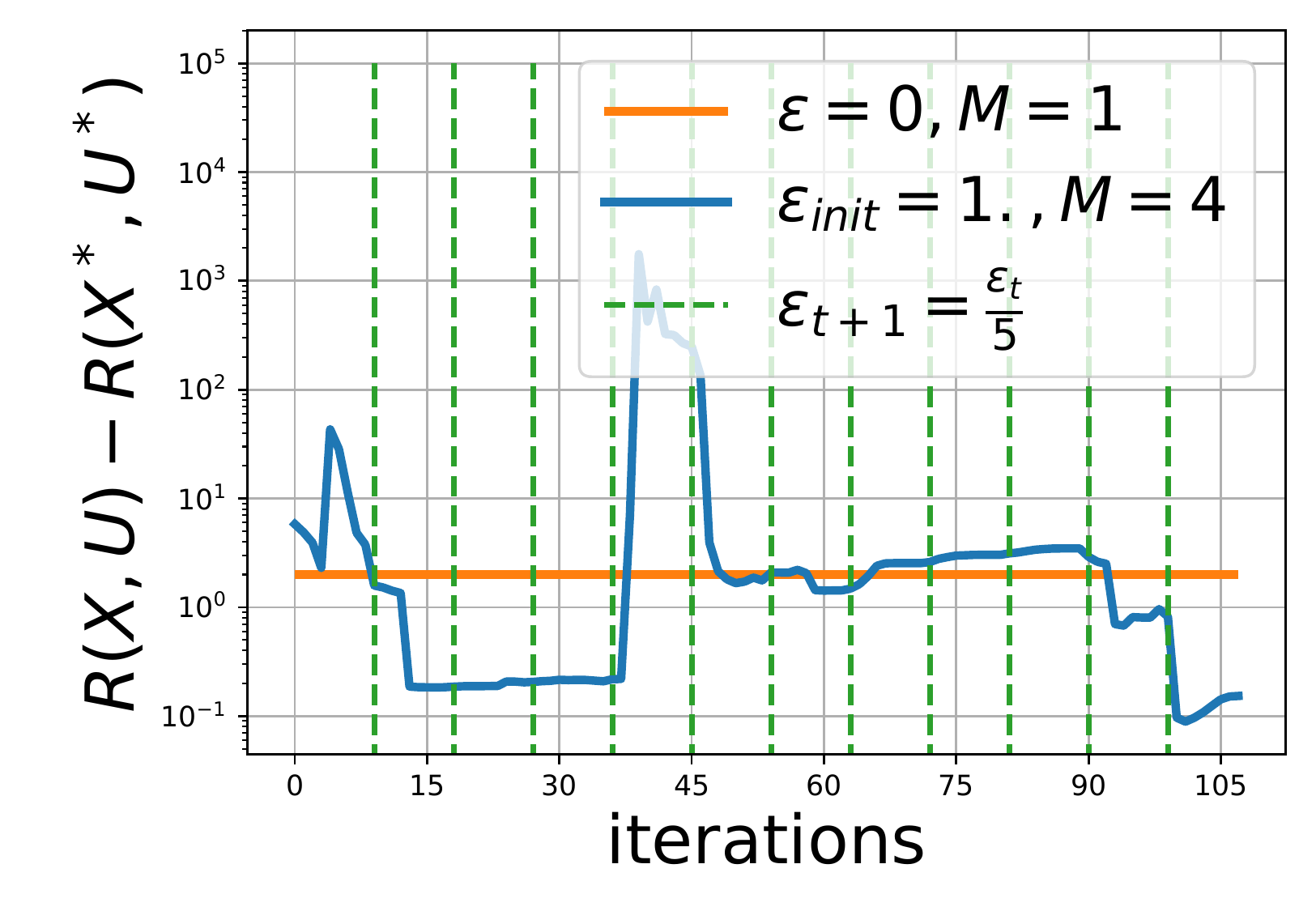}
    \includegraphics[width=0.45 \linewidth]{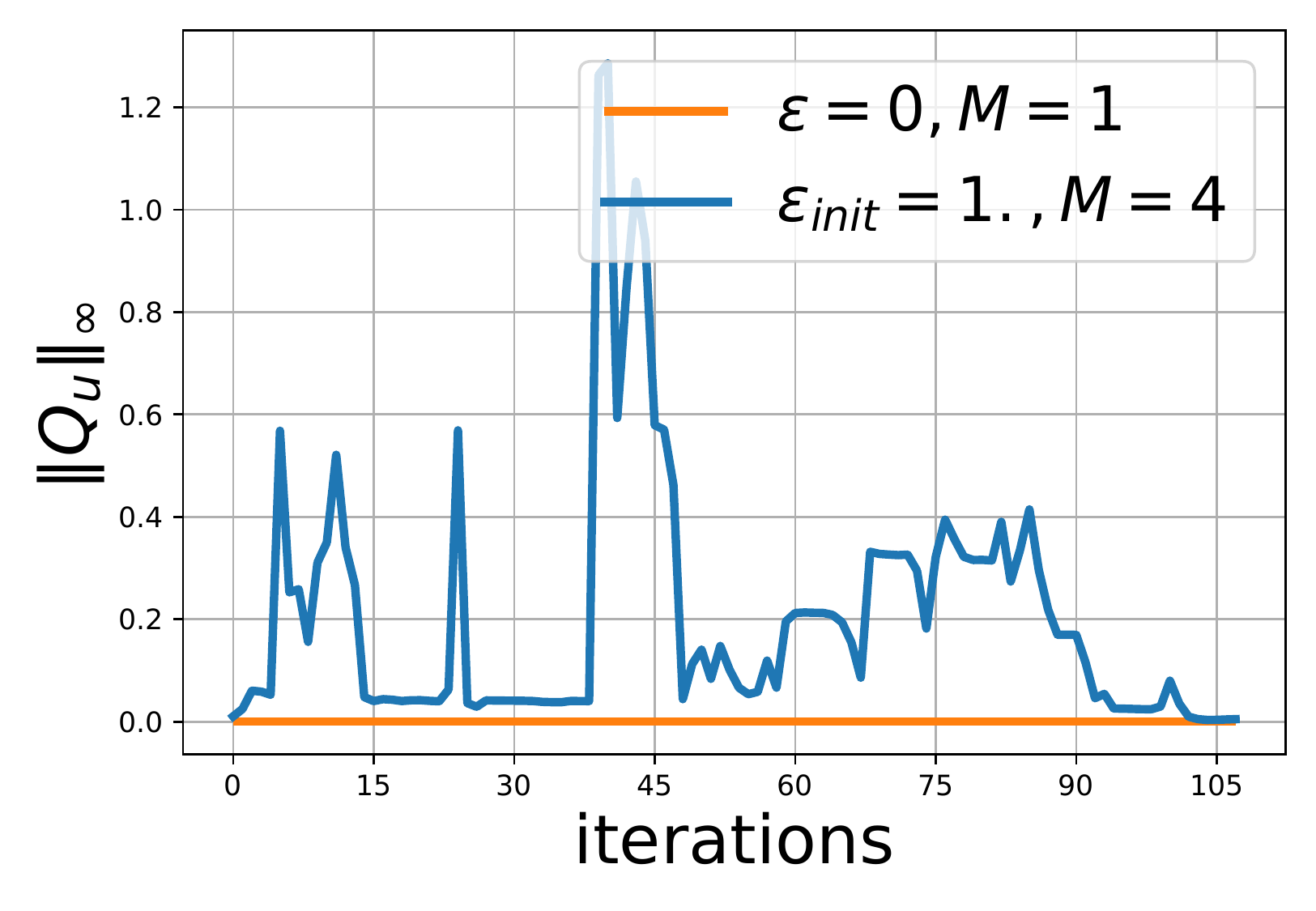}
    \caption{RDDP allows executing a single-legged jump (on a vertical plane) towards a target position marked with a blue sphere. \textbf{Top:} The leg initially bends downwards and then extends to perform the jumping motion. \textbf{Bottom:} The cost $J$ and the norm $\|Q_u\|$ are constant when using classical DDP (orange) while RSDDP (blue) achieves to lower the cost. The vertical green lines denote updates in the noise level used in randomized smoothing.}
    \label{fig:pictures_solo_leg}
    \vspace{-0.0cm}
\end{figure*}

To illustrate the issues induced by the non-smoothness of unilateral contact and frictions, we consider the preliminary tasks of taking off or sliding a cube on a table (appearing in \cite{werling2021fast}). 
In these experiments, we use a cost function similar to the one of \eqref{eq:cost_pendulum}, optimizing the difference between a target configuration and the final configuration of the system at the end of the planned trajectory.
As shown by Fig.~\ref{fig:cube_double_cartpole}~(left), our approach allows us to complete this task while we have shown in Sec.~\ref{sec:local_OC} that it is impossible when relying on classic gradient information.

Similarly, it is possible to apply the RDDP algorithm to control more complex systems such as a double pendulum and a cartpole with dry friction on the joints (Fig.~\ref{fig:cartpole}).
The double pendulum and the cartpole are illustrative instances of underactuated systems whose dynamics are non-smooth due to friction.
In both cases, classical DDP fails to optimize the control variable using the true dynamics.
Once again, smoothing the dynamics with respect to the control $u$ using $M$ samples per time step allows to have non-null gradients almost everywhere and thus to apply classical DDP (Fig.~\ref{fig:cube_double_cartpole}, middle and right).
Combining randomized smoothing with the DDP algorithm makes it possible to solve the problem of controlling a complex system involving contacts and friction very efficiently. 
Interestingly, the FDDP algorithm \cite{mastalli2020crocoddyl}, which allows for unfeasible trajectories, also failed to overcome the non-smoothness from dry frictions.
We also studied the influence of the number of samples $M$ used for the MC estimators (Fig.~\ref{fig:cartpole}) and the adaptive scheme (Fig.~\ref{fig:cube_double_cartpole}) on the quality of the obtained solution. 
We noticed that good results could be obtained with a relatively small number of samples $M$ for the first-order MC gradient estimator, meaning that the supplementary computational costs are limited when compared to classical DDP. 
On the contrary, using an adaptive scheme for decreasing the smoothing perturbation leads to a more precise solution by allowing the gradients of the trajectory optimization problems to converge to zero (Fig.~\ref{fig:cube_double_cartpole}).
This represents an advantage of our method when compared to classical RL approaches, as the latter requires keeping a non-null noise to estimate gradients, preventing them from converging very precisely.

\subsection{Controlling contact interactions in robotics}
The following experiments explore the application of our method to control complex robotics systems interacting with their environments via frictional contacts.\\

\noindent
\textbf{Quadrotor.} In our first experiment, we control the takeoff of a quadrotor.
The drone starts in an initial state, landing on the ground, and the goal is to reach a target position one meter higher in the air.
This case of a drone with four actuated thrusts represents a realistic and challenging robotics task as one could need to control its trajectory without having to plan for takeoff and landing phases, which would require breaking or making unilateral contacts.
Concretely,  the cost of the control problem $J$ is similar to \eqref{eq:cost_pendulum} but with $p$ now denoting the position of the base of the quadrotor with $w_p = 4$ and $w_u = 10^{-3}$.
As shown by Fig.~\ref{fig:pictures_quad}, the classical DDP algorithm fails due to the issue of non-informative gradients whenever the drone rests on the ground.
In contrast, our approach is quickly (less than 10 optimization steps) able to generate a trajectory that accurately reaches the target position.\\

\noindent
\textbf{Solo robot.}
Finally, we apply our algorithm to solve a task on a legged robotic system with frictional contacts. 
We use one leg of the Solo robot~\cite{grimminger2020open} similar to \cite{viereck18hopper}, which results in a hopper with 3 degrees of freedom, namely the hip and knee joint, and a vertical translation of the base. Therefore, the state is fully described by the position $q \in \mathbb{R}^3$ and velocity $\dot{q} \in \mathbb{R}^3$. The system is underactuated as only the knee and the hip of the robot are actuated. 
The task is to make the system jump up from an initial stretched configuration (Fig.~\ref{fig:pictures_solo_leg}, \textbf{Left}) to a position 30cm above the ground along the vertical axis using a similar formulation for the control cost $R$ as in \eqref{eq:cost_pendulum}.
Without randomized smoothing, the DDP algorithm has no gradient information in this configuration; the algorithm converges to a local optimum, and the system is not moving. 
On the contrary, using RDDP achieves solving the task by first bending the leg downward (Fig.~\ref{fig:pictures_solo_leg}, $3^{\text{rd}}$ image), to be then able to apply a control torque that leads to contact forces with the ground moving the leg up to the target position.
It is worth noting that this movement was generated without pre-specifying contact phases, as the optimizer does the contact planning entirely.
This represents an encouraging first step towards automatically generating locomotion movements via trajectory optimization.

\section{Conclusion}
Analyzing reinforcement learning via the theory of randomized smoothing allows us to understand how crucial the exploratory characteristic of these algorithms is to solve control problems with non-smooth dynamics.
By transferring these ideas to the field of trajectory optimization, we have leveraged randomized smoothing in this paper to propose an approximate and smooth formulation of the original optimal control problem.
Exploiting this new formulation with the well-established DDP algorithm results in an approach that can cope with the presence of contact interactions and friction in a sample-efficient manner. 
We've also demonstrated the capacity of our method to properly cope with standard robotics systems (pendulum, cart pole, drone, Solo robot), including non-smooth dynamical effects (frictions, contacts) where classic optimization-based are likely to fail.
In future work, we plan to investigate possible uses of the information contained in the variance of the several particles of the Monte-Carlo estimator. 
In particular, we believe this could help to build a more robust adaptive scheme by detecting when increasing the noise $\epsilon$, the precision threshold $\alpha$, or even the number of particles $M$ is necessary.

\section*{Acknowledgements}
We warmly thank Robin Strudel for the fruitful discussions and feedback on the paper.
This work was supported in part by L'Agence d'Innovation Défense, the French government under the management of Agence Nationale de la Recherche through the projects INEXACT (ANR-22-CE33-0007-01) and NIMBLE (ANR-22-CE33-0008), as part of the "Investissements d'avenir" program, reference ANR-19-P3IA-0001 (PRAIRIE 3IA Institute), and by the  European Union through the AGIMUS project (GA no.101070165) and the Louis Vuitton ENS Chair on Artificial Intelligence.

\balance
\bibliographystyle{IEEEtran}
\bibliography{IEEEabrv,references}

\appendix
\section*{Appendix}\label{sec:app-ncp}
This section presents the physical principles commonly used for rigid body simulation with point contact to show where non-smoothness and non-convexity arise during the computation of $f$ in \eqref{eq:dyn_cons}.
In general, we describe the state of a system with its generalized coordinates $q \in \mathcal{Q} \cong \mathbb{R}^{n_q}$ and denote by $v \in \mathcal{T}_q\mathcal{Q} = \mathbb{R}^{n_v}$ the joint velocity.
The principle of least action states that the path followed by a dynamical system should minimize the action functional which induces the
Lagrangian equations of motion:
\begin{equation}\label{eqn:app-1}
    M(q) \dot{v} + C(q, v)v + g(q) = \tau
\end{equation}
where $M \in \mathbb{R}^{n_v \times n_v}$ represents the joint space inertia matrix of the system, $C(q, v)v$ accounts for the centrifugal and Coriolis effects, and $g$ is the generalized gravity.
In the following, we express the problem in terms of velocities rather than acceleration, thus discretizing \eqref{eqn:app-1} to
\begin{equation}\label{eqn:app-2}
    Mv_{t+1} = Mv_t + (\tau - Cv - g)\Delta t
\end{equation}
where $C$ and $g$ are evaluated explicitly at $(q_t,v_t)$
We note $v_f$ the free velocity which is defined as the solution of \eqref{eqn:app-2}.

\noindent
\textbf{Constrained dynamics}
for robotic systems with kinematic loops or that establish contacts with the environment can define these constraints implicitly via
\begin{equation}\label{eqn:app-3}
    \Phi(q) = 0
\end{equation}
where $\Phi : \mathbb{R}^{n_q} \rightarrow \mathbb{R}^m$ is the implicit constraint function of dimension $m$ depending on the nature of the constraint.
\eqref{eqn:app-3} is then derived w.r.t time to express the constraint as a constraint on joint velocities in the sequel:
\begin{equation}\label{eqn:app-4}
    c - c^* = 0
\end{equation}
where $c = Jv_{t+1}$ is the constraint velocity, $J = \frac{\partial\Phi}{\partial q}$ is the constraint Jacobian and $c^*$ is the reference velocity of the constraint which is set to either model physical effects or stabilize the numerics.
Such a constraint is enforced by the action of the environment on the system via the contact vector impulse $\lambda \in \mathbb{R}^m$ and should be incorporated into the Lagrangian equations of motion~\eqref{eqn:app-1}:
\begin{equation}\label{eqn:app-5}
    Mv_{t+1} = Mv_f + J^T\lambda
\end{equation}

\noindent
\textbf{Unilateral contacts} are usually used when a system is in contact with its environment and enforce the normal component of the separation vector, i.e. the signed distance function, to be non-negative:
\begin{equation}\label{eqn:app-6}
    \Phi(q)_N \geq 0
\end{equation}
where $\Phi(q) \in \mathbb{R}^{3 \times n_c}$, $n_c$ is the number of contacts, and the subscripts $N$ and $T$ respectively account for the normal and tangential components. Similarly to \eqref{eqn:app-4}, this can be expressed as:
\begin{equation}\label{eqn:app-7}
    c_N - c^*_N \geq 0
\end{equation}
where $c = Jv_{t+1} \in \mathbb{R}^{3 \times n_c}$ is the velocity of contact points and $c^*_N$ is set to model physical effects or improve the numerical accuracy of the solutions (commonly referred to as Baumgarte stabilization).
Unilateral contacts constrain the possible efforts $\lambda$ depending on the state of the system and the friction coefficient. In general, the contact forces $\lambda$ can only be repulsive as they should not act in a glue-like fashion (the environment can only push and not pull on the feet of a legged robot) and, thus, are forced to be non-negative. An impulse cannot occur when an object takes off, implying that the normal velocity and impulse cannot be non-null simultaneously.

In a frictionless situation, the tangential forces are null, which implies that $\lambda_T = 0$.  Combining these conditions, we obtain the so-called Signorini condition:
\begin{equation}
    0 \leq \lambda_N \perp c_N - c^*_N \geq 0
\end{equation}
However, such a condition does not define a mapping between $\lambda_N$ and $c_N$, i.e., the contact forces are not a function of the penetration error. Indeed, its representation is an infinitely steep graph and this problem is referred to as a Linear Complementarity Problem (LCP).
Therefore, the computation of $f$ in \eqref{eq:dyn_cons} which corresponds to the computation of $v_{t+1}$ in \eqref{eqn:app-5} and $q_{t+1}$ via integration is facing the issues of non-smoothness and non-convexity.

If friction is included, additional conditions on $\lambda_T$, i.e. lying in a friction cone and satisfying the maximum dissipation principle, lead to a reformulation of the problem as a nonlinear complementarity problem (NCP), see \cite{lidec2023contact} for further details.

\end{document}